\documentclass[journal]{IEEEtran}

\usepackage[ruled,vlined,algo2e]{algorithm2e}
\usepackage{cite}
\usepackage{epsfig}
\usepackage{epstopdf}
\usepackage{cite}
\usepackage{array,multirow}
\usepackage{makecell}
\usepackage{subfigure}
\usepackage{color, colortbl}
\usepackage[table]{xcolor}
\usepackage{amssymb}
\usepackage{bbding}
\usepackage{pifont}
\usepackage{amsfonts}
\usepackage{amsmath}
\usepackage[Code]{algorithm}
\usepackage {algorithmic}
\usepackage{threeparttable}
\usepackage{color, colortbl}
\usepackage{stfloats}
\usepackage{mathrsfs,amsmath}

\usepackage{amsmath}
\usepackage{amssymb}

\usepackage{tabularx}
\usepackage{array}

\usepackage{makecell}

\usepackage{graphicx}
\usepackage{color}

\usepackage[normalem]{ulem}

\usepackage{hyperref}
\usepackage{diagbox}

\begin{document}
\title{FPGA-based Acceleration for Convolutional Neural Networks: A Comprehensive Review}

\author{
        Junye~Jiang\IEEEauthorrefmark{2},
        Yaan~Zhou\IEEEauthorrefmark{2},
        Yuanhao~Gong\IEEEauthorrefmark{2},
        Haoxuan~Yuan,
        and~Shuanglong~Liu\IEEEauthorrefmark{1}
        
\thanks{This work was supported in part by the Scientific Research Foundation of Hunan Provincial Education Department (Key project 23A0087) and Changsha Municipal Natural Science Foundation (No. kq2502001). }

\thanks{Junye Jiang, Yaan Zhou, Yuanhao Gong, Haoxuan Yuan and Shuanglong Liu are with the Key Laboratory of Low-Dimensional Quantum Structures and Quantum Control of Ministry of Education, School of Physics and Electronics, Hunan Normal University, Changsha, 410081 China.} 

\thanks{\IEEEauthorrefmark{2} Equal Contribution.}

\thanks{\IEEEauthorrefmark{1} Corresponding author: Shuanglong Liu (liu.shuanglong@hunnu.edu.cn).}

}


\maketitle

\begin{abstract}

Convolutional Neural Networks (CNNs) are fundamental to deep learning, driving applications across various domains. However, their growing complexity has significantly increased computational demands, necessitating efficient hardware accelerators. Field-Programmable Gate Arrays (FPGAs) have emerged as a leading solution, offering reconfigurability, parallelism, and energy efficiency.
This paper provides a comprehensive review of FPGA-based hardware accelerators specifically designed for CNNs. 
It presents and summarizes the performance evaluation framework grounded in existing studies and explores key optimization strategies, such as parallel computing, dataflow optimization, and hardware-software co-design. 
It also compares various FPGA architectures in terms of latency, throughput, compute efficiency, power consumption, and resource utilization. Finally, the paper highlights future challenges and opportunities, emphasizing the potential for continued innovation in this field.

\end{abstract}





\IEEEpeerreviewmaketitle

\section{Introduction}
\label{sec:introduction}

In recent years, Deep Learning (DL) has gained significant attention in computer science \cite{chauhan2024brief}, with Convolutional Neural Networks (CNNs) serving as a cornerstone of this progress. CNNs have achieved remarkable success across various applications, including image classification \cite{chen2021review}, object detection \cite{ashiq2022cnn}, and semantic segmentation \cite{gidaris2015object}. Their transformative impact has led to widespread adoption by industry leaders, recognizing CNNs as a vital tool for innovation and competitive advantage.

The strength of CNNs lies in their ability to process high-dimensional data through convolutional operations, pooling, and non-linear activations, enabling automated multi-level feature extraction. However, this capability comes at the cost of increased computational complexity and resource demands. With their adoption in mobile and edge devices like drones and smartphones \cite{wang2024repvit}, CNNs face critical challenges in achieving high performance under resource constraints.

To address these challenges, extensive research has been conducted on hardware acceleration for CNNs \cite{liu2021toward, dai2024dcp, hegde2018morph, moolchandani2021accelerating}. While CPUs, GPUs and ASICs are commonly employed, their limitations in handling the complexity of modern CNN architectures and real-time processing needs have prompted growing interest in alternative platforms. Field-Programmable Gate Arrays (FPGAs) have emerged as a promising solution, offering reconfigurability, parallel processing, and energy efficiency. Their ability to customize hardware for specific tasks enables the design of high-performance, resource-efficient accelerators tailored for deep learning applications.

This review article explores recent advances in FPGA-based CNN accelerators, focusing on acceleration methods, architectural innovations, hardware optimization techniques, and software-hardware co-design frameworks. 
It highlights key trends, addresses current challenges, and outlines future directions in this rapidly evolving domain.
Additionally, the article examines performance evaluation metrics and provides insights into diverse architectures, offering practical guidance for researchers and practitioners.
The main contributions of this work are summarized as follows:

\begin{itemize}

\item	\textbf{Evaluation Metrics}: We review FPGA-based CNN acceleration research, consolidating key performance metrics, particularly compute and overall efficiency. It explores the relationships between these metrics, aiding researchers in identifying design limitations and guiding future advancements (Section~\ref{sec:metrics});

\item	\textbf{Acceleration Methods}: We present a quantitative comparison of existing CNN acceleration methods, providing practitioners and researchers with valuable insights to select approaches that best align with their specific system requirements (Section~\ref{sec:methods});

\item	\textbf{Parallel Computing Analysis}: This study reviews parallel computing methods and their hardware architectures in existing designs, analyzing their strengths and weaknesses using the introduced evaluation metrics. It highlights the importance of dynamic parallelism in enhancing hardware performance, including compute efficiency and resource utilization (Section~\ref{sec:hardware});

\item   \textbf{Future Directions}: Building on the comprehensive review in this work, we propose and discuss future research directions to help researchers and designers identify key design considerations and potential optimizations in this field (Section~\ref{sec:discuss}).

\end{itemize}

The rest part of this paper is organized as follows:
Section~\ref{sec:platform} compares the acceleration platforms for CNNs, emphasizing the advantages of FPGAs.
In Section \ref{sec:metrics}, we analyze and outline design performance metrics for FPGA-based CNN accelerators, guiding the comparison of existing designs in subsequent sections. This section also highlights commonly overlooked metrics in the literature and suggests directions for future design and optimization.
Section \ref{sec:methods} presents acceleration methods for CNNs at both the algorithmic and hardware levels, with qualitative and quantitative comparisons.
Section \ref{sec:hardware} reviews existing FPGA-based CNN accelerators, focusing on parallel computing techniques in hardware design.
Section~\ref{sec:co-design} discusses hardware-software co-design methodologies, including toolflows, design space exploration, and performance and resource modeling.
Section~\ref{sec:discuss} concludes this paper with a discussion on future research directions.

\section{CNN Acceleration Platforms}
\label{sec:platform}

This section compares the architectural and performance characteristics of common hardware acceleration platforms, including central processing units (CPUs), graphics processing units (GPUs), application-specific integrated circuits (ASICs), and field-programmable gate arrays (FPGAs). It focuses on their performance in CNN applications, emphasizing that FPGA-based accelerators deliver superior performance for deep learning tasks.

\subsection{Hardware Platforms}

\subsubsection*{\textbf{Central Processing Units (CPUs)}}

CPUs are fundamental components in modern computing systems, adhering to the von Neumann architecture. In the context of CNN models, CPUs are well-suited for handling complex logical operations, such as model initialization, data preprocessing, and managing communication between hardware components. They excel in processing diverse data types and executing intricate computations independently. However, their limited capacity for parallel processing makes them less efficient for the high-volume matrix operations and convolutions characteristic of CNNs. Additionally, CPUs often face challenges like high power consumption and restricted memory bandwidth, which can become bottlenecks when processing the large-scale data and computations required by CNN models \cite{gyawali2023comparative}.

\subsubsection*{\textbf{Graphics Processing Units (GPUs)}}

GPUs have been rapidly developed to handle large-scale computations, particularly in applications like image processing and CNN models. Unlike CPUs, GPUs are optimized for parallel computing, capable of processing massive amounts of similar data simultaneously, making them well-suited for tasks like CNN training, where multiple convolutions and matrix operations can be executed in parallel. The architecture of GPUs, with larger arithmetic logic units, enables faster processing speeds and enhances their parallel computing capabilities.

For example, in CNN-based applications, GPUs significantly speed up tasks such as training deep learning models by accelerating the computation of convolutional layers and weight updates \cite{zhang2019recent}. However, GPUs have limitations in single-threaded performance, which makes them less ideal for tasks requiring complex logic or single-core execution. Additionally, GPUs face challenges such as high power consumption, substantial cooling needs, and limited memory capacity when dealing with extremely large datasets, which can hinder performance in ultra-large-scale CNN models.

\subsubsection*{\textbf{Application-Specific Integrated Circuits (ASICs)}}

ASICs are specialized integrated circuits designed to meet specific product requirements. The main advantage of ASICs  in deep learning applications lies in their customization, which allows for the design of high-performance circuits tailored to the specific operations of CNNs, such as convolution, activation, and pooling layers. 
This customization allows for the elimination of redundant components, significantly reducing chip area and power consumption compared to CPUs and GPUs \cite{beldianu2014ASIC}. 

However, the highly specialized nature of ASICs comes with trade-offs. The design process is time-consuming, and ASICs are highly dependent on the particular CNN models they are designed for, making them less adaptable to evolving architectures or new tasks. Furthermore, ASICs lack the flexibility to rapidly adjust to the dynamic nature of CNN development, which limits their applicability as deep learning models evolve. Unlike FPGAs, which can be reprogrammed or dynamically configured, ASICs are optimized for specific tasks, making them less versatile and unsuitable for handling diverse workloads outside their narrow focus.

\subsubsection*{\textbf{Field Programmable Gate Array (FPGAs)}}

To enhance the programmability of hardware modules, FPGAs are increasingly utilized in various applications, thanks to their flexibility and reconfigurability. In the context of CNNs, FPGAs serve as a low-latency hardware platform that allows for customized data path designs and optimized memory hierarchies, making them a valuable complement to CPUs and GPUs. FPGAs excel in maximizing performance through dynamic task partitioning and efficient software-hardware co-design, which are critical for optimizing CNN operations. These features make FPGAs particularly advantageous in AI inference tasks, especially in resource-constrained environments like edge computing, where power efficiency and flexibility are paramount \cite{jose2024training}.
For this reason, this article provides a detailed literature review of FPGA-based CNN accelerators, which will be discussed in the following sections.

\subsection{Further Comparison of FPGAs and GPUs}

The acceleration of CNNs consists of two main stages: training and inference.
CNN training involves constructing and optimizing the model through iterative processes across multiple epochs, refining the model parameters with each cycle.
CNN inference uses a trained model to make predictions, such as classification or detection, without the need for weight updates or gradient calculations.

In the training phase, GPUs excel in leveraging their high parallelism, achieving substantial throughput by processing large batches of data. However, their efficiency depends heavily on batch size, which works well in offline scenarios but is less effective in real-time processing, particularly for CNN models requiring continuous data transfer. In such cases, GPU performance tends to degrade \cite{guo2017angel}. Additionally, GPUs often introduce significant latency, especially during convolution operations, which can impact overall system performance.

On the other hand, dedicated architectures for CNNs can be designed to parallelize computations within a frame. The flexibility of FPGAs makes them a promising candidate for CNN acceleration. With a scalable design, CNN accelerators can also be implemented on embedded FPGAs.
While GPUs offer inherent parallelism, FPGAs allow for finer-grained parallelism by customizing the hardware architecture to optimize specific CNN layers. This customization enables faster execution of operations, such as convolution and activation, thereby improving overall CNN performance \cite{liu2019towards}.

\subsection{Discussion}

\begin{table*}[t]
\centering

\caption{Comparison of CNN Acceleration Platforms}

    \renewcommand\arraystretch{1.5}
    \resizebox{0.85\textwidth}{!}
{  
{ \begin{tabular}{|m{0.15\textwidth}<{\centering}|m{0.15\textwidth}<{\centering}|m{0.15\textwidth}<{\centering}|m{0.15\textwidth}<{\centering}|m{0.15\textwidth}<{\centering}|} \hline 

Metrics &  CPUs &  GPUs &  ASICs & FPGAs  \\ \hline 
         
Compute Performance & 
  \ding{56}\ding{56} Limited computing capability with general-purpose &  \ding{52}\ding{52}\ding{52} Highly parallel computing capability &   \ding{52}\ding{52} Highly parallelized for specific applications &  \ding{52}\ding{52} Highly parallelized with customizable hardware 
  \\ \hline 
         
Energy Efficiency  &  \ding{56} Moderate power &  
\ding{56}\ding{56} Large power &  
\ding{52}\ding{52}\ding{52} Lowest Power &  
\ding{52}\ding{52} Low power  \\ \hline

Flexibility &  \ding{52}\ding{52} Highly flexible &  
\ding{52} Limited flexibility &
\ding{56} Fully customized &
\ding{52}\ding{52} Programmable \\ \hline 

Development Time &  \ding{52}\ding{52}\ding{52} Shortest development time &  \ding{52}\ding{52} Moderate development time &
\ding{56}\ding{56}\ding{56} Time-consuming development & 
\ding{52} Acceptable development time \\ \hline 

Cost &  \ding{56}\ding{56} High cost & 
\ding{56}\ding{56}\ding{56} Largest cost &  
\ding{52} High initial cost but cost-effective for large-scale production & \ding{52} Relatively low \\ \hline 


Model Scalability &  \ding{52}\ding{52} Reconfigurable &
\ding{52}\ding{52} Reconfigurable & 
\ding{56}\ding{56} Worst scalability &
\ding{52} Scaled with reconfiguration \\ \hline

    \end{tabular}
    }
 } 
\vspace{-2ex}
\label{tab:my_label}
\end{table*}

Table \ref{tab:my_label} summarizes the performance comparison of CPUs, GPUs, ASICs, and FPGAs across various metrics such as computing performance, energy efficiency, flexibility, development time, cost, and scalability. Based on these factors, a comparative analysis of the platforms is provided.

FPGAs stand out in accelerating CNN models due to their customization and flexibility. Unlike ASICs, which are fixed to specific tasks, FPGAs can be reprogrammed to optimize performance for different CNN architectures, making them ideal for applications requiring a balance between accuracy and throughput. Additionally, FPGAs can leverage sparsity techniques to significantly enhance throughput. For example, the sparsity-aware CNN accelerator by Wen \textit{et al.} \cite{wen2020kernel} achieved up to 89.7\% higher throughput compared to baseline designs.

FPGAs also offer advantages in power efficiency, particularly for edge devices with strict power constraints, providing a good balance between flexibility and low power consumption. While ASICs may offer lower power usage, FPGAs are more adaptable and power-efficient than GPUs and CPUs. Furthermore, FPGAs excel in reducing latency, making them ideal for real-time applications that demand quick responses~\cite{Nurvitadhi2016int}.
\section{Evaluation Framework}
\label{sec:metrics}

In CNN acceleration, evaluation metrics are crucial for assessing design performance. These metrics fall into two categories: network model performance and hardware accelerator evaluation. This section focuses on the key metrics used to evaluate FPGA-based CNN accelerators, highlighting throughput, compute efficiency, and energy efficiency as critical factors for hardware optimization.

\subsection{Model Performance Evaluation}

\subsubsection*{\textbf{Accuracy}}

Algorithm adaptation and optimization techniques, such as model compression and data quantization, are often employed to reduce complexity. As a result, evaluating the accuracy of optimized models becomes crucial.
Accuracy is a widely used metric in CNN model design to assess performance, representing the proportion of correctly classified samples in a dataset. It is calculated using the formula:
\begin{equation} 
\text{Accuracy} = \frac{TP+TN}{TP+TN+FP+FN} 
\end{equation} 
where $TP$ (True Positive), $TN$ (True Negative), $FP$ (False Positive), and $FN$ (False Negative) indicate the model's correct and incorrect predictions. Accuracy provides an overall measure of classification performance, especially when the dataset has a balanced distribution of positive and negative samples. In such cases, a high accuracy rate typically indicates better performance. However, accuracy has limitations, particularly in the presence of class imbalance. In these cases, a high accuracy rate may mask poor performance on minority classes.

\subsubsection*{\textbf{Precision and Recall}}

To address the limitations of accuracy, it is often used in conjunction with precision and recall~\cite{buckland1994relationship}.
Precision measures the model's accuracy in predicting positive classes, indicating the proportion of predicted positive samples that are correctly identified as positive. It is calculated as: 
\begin{equation} \text{Precision} = \frac{TP}{TP+FP} \end{equation}

However, precision does not consider false negatives and focuses only on the accuracy of positive predictions, ignoring cases where the model fails to identify actual positive samples. In scenarios where missing positive instances has a high cost, relying solely on precision may not be enough. To address this, Recall is used to evaluate the model's ability to correctly identify all positive samples. Recall is calculated using the formula: 
\begin{equation} \text{Recall} = \frac{TP}{TP+FN} \end{equation}

\subsubsection*{\textbf{F1 Score}}

Precision and recall often exhibit an inverse trade-off. Increasing recall typically leads to a decrease in precision, and vice versa. This occurs because when the model prioritizes predicting positive classes to improve recall, the number of false positives may rise, resulting in lower precision. On the other hand, if the model overly emphasizes precision, it might miss some true positive samples, leading to a reduction in recall.

To balance this trade-off, the F1 score is commonly used~\cite{yacouby2020probabilistic}. It is the harmonic mean of precision and recall, offering a comprehensive evaluation of model performance across both metrics. It is calculated as follows: 
\begin{equation} \text{F1} = 2 \times \frac{\text{Precision} \times \text{Recall}}{\text{Precision} + \text{Recall}} \end{equation}

\subsection{Hardware Performance Evaluation}

Various hardware performance metrics, including throughput, latency, and power consumption, are widely used to evaluate and compare FPGA-based CNN accelerators. However, achieving a fair comparison requires careful consideration of the operations or workloads of the CNN model, the available FPGA resources, and the underlying chip technology. Here, we provide a detailed discussion of the commonly employed evaluation metrics in the literature, highlighting their significance and impact.

\subsubsection*{\textbf{Latency}}

Latency is a key consideration for CNN hardware accelerators, representing the runtime taken from data input to the completion of processing and output generation. 
It serves as a critical measure of the system's responsiveness when processing a single image.
Low latency is especially important for real-time applications such as autonomous driving \cite{li2022traffic} and speech recognition \cite{hema2023emotional}. The overall latency of a design can be divided into three primary components: (1) Off-chip data transfer latency, determined by the model parameters and the memory bandwidth; (2) On-chip data transfer latency, influenced by the efficiency of internal data routing within the hardware; (3) Computation latency, dependent on the number of operations required by the CNN model and the level of parallelism implemented in the hardware design.

To minimize latency in CNN accelerators, the following optimization strategies are commonly employed:

\begin{itemize}
    \item Increasing memory bandwidth: Utilize high-bandwidth, low-latency memory technologies such as High Bandwidth Memory (HBM) to accelerate data read and write processes \cite{jiang2021optimized};
    
    \item Reducing data transfer bottlenecks: Employ efficient data transfer protocols or advanced bus architectures, such as PCIe \cite{wan2024pflow} or NVLink \cite{nabavinejad2020overview}, to lower latency between accelerators and external devices;
    
    \item Optimizing data storage patterns: Enhance data reuse and reduce the frequency of data transfers by employing optimized storage patterns \cite{li2022efficient}, thereby decreasing data transfer latency;
    
    \item Enhancing computational parallelism: Design hardware architectures to maximize computational parallelism, enabling faster processing within the accelerator and reducing overall computational latency \cite{dai2024dcp}.

\end{itemize}

\subsubsection*{\textbf{Throughput}}
\label{sec:througput}

Throughput, often referred to as hardware performance, measures the number of computational operations a system can process within a specific time frame. Unlike latency, throughput accounts for the computational workloads of different CNN models. As such, it serves as a key metric for evaluating hardware efficiency and performance \cite{huang2022computer}. The relationship between CNN workloads, throughput, and latency can be expressed as:
\begin{equation}
\text{Throughput} = \frac{\text{Workloads}}{\text{Latency}}
\end{equation}
Here, workloads refer to the volume of computational operations in the CNN model.

It is important to differentiate system throughput from the theoretical peak performance of the design, which is defined as:
\begin{equation}
\label{eq:Trt}
\text{{Peak}\_{throughput}\_{design}} = 2 \times \#\text{MACs} \times f
\end{equation}
where $\#\text{MACs}$ represents the number of multiply-accumulate (MAC) units in the hardware accelerator, and $f$ denotes the operating clock frequency of the design. This formula represents the theoretical maximum performance of the realized hardware architecture.

While the theoretical peak throughput indicates the maximum achievable performance of the hardware design, actual throughput reflects the real-world performance. According to Eq. (\ref{eq:Trt}), the peak throughput is influenced by the number of MAC units, which is directly related to the parallelism of the hardware framework. Larger parallelism leads to more MAC units and increased computational capacity.

\subsubsection*{\textbf{Compute Efficiency}}

The theoretical peak throughput represents the ideal maximum performance of a hardware system, acting as the upper limit of an accelerator’s capacity. However, in practical CNN acceleration, the network parameters may not align with the computational parallelism of the design. This misalignment can cause unbalanced workloads among the processing elements (PEs) in the computation engine, leading to suboptimal utilization of available resources.
Additionally, factors such as memory bandwidth and data transfer rates significantly impact throughput, often creating bottlenecks. As a result, actual throughput is lower than the theoretical peak performance due to these practical constraints.

To evaluate the efficiency of a hardware design, compute efficiency, also referred to as MAC efficiency, is employed~\cite{wu2019compute}. This metric is defined as the ratio of the actual throughput to the peak theoretical throughput. It serves as a critical measure of hardware accelerators' effectiveness by quantifying the proportion of useful MAC cycles relative to the total MAC units in the design. Compute efficiency is calculated using the formula:
\begin{equation}
\text{Compute  EFF.} = \frac{\text{Throughput}}{2 \times \#\text{MACs} \times f}
\end{equation}

Unlike other evaluation metrics mentioned above, compute efficiency is independent of the specific FPGA device or the number of MAC units utilized. Instead, it reflects the intrinsic efficiency of the accelerator design, making it a more reliable measure for assessing and comparing different hardware architectures for CNN acceleration.

\subsubsection*{\textbf{Resource Efficiency}}

In CNN accelerators, resource efficiency measures the system's throughput relative to the number of DSP units used \cite{fan2021high}. This metric evaluates the contribution of each DSP unit to the computational task and is defined as:
\begin{equation}
\text{Res. EFF.} = \frac{ \text{Throughput} }{\text{DSP\_{used}}}
\end{equation}

Higher resource efficiency reflects better DSP utilization, with each unit delivering significant computational throughput. This enhances hardware resource usage while maintaining high performance, reducing power consumption and idle resources, and improving overall system efficiency and stability.
However, in designs where logic resources are used alongside DSPs to implement MACs, this metric may not provide a fair basis for comparison across different designs.

\subsubsection*{\textbf{Clock Efficiency}}

Clock efficiency measures how effectively a hardware accelerator operates relative to its maximum clock frequency \cite{jiang2023automated}. It is defined as:
\begin{equation}
\text{Clock EFF.} = \frac{ \text{Working Clock} }{ \text{Maximum Clock} }
\end{equation}

This ratio indicates the utilization of the hardware's computational potential. Higher clock efficiency means the accelerator operates near its peak frequency, while lower efficiency suggests underutilization. Practical constraints like power consumption, heat dissipation, and system stability often limit clock efficiency.
Improving heat dissipation, power management, and task scheduling can boost clock efficiency, leading to better performance in high-demand applications such as real-time processing and data centers.

\subsubsection*{\textbf{Overall Efficiency}}

Corresponds to the peak throughput of the design mentioned above, the theoretical peak performance of an FPGA device is calculated by multiplying the total number of multipliers and adders in its DSP blocks by the maximum clock rate \cite{parker2014understanding}. This value represents the theoretical computational limit, which is unattainable in practice because real-world algorithms cannot continuously utilize all computational units. However, it serves as a valuable benchmark for comparison.

The overall efficiency is defined as the ratio of achieved performance to the peak performance of the device, offering a comprehensive measure of the hardware's effectiveness \cite{liu2021toward}. It can be expressed as:
\begin{equation}
  \textsc{Overall Eff.} = \textsc{Clock Eff.}\times \textsc{Res. Uti.}\times \textsc{Compute Eff.}
\end{equation}
where clock efficiency is the ratio of the working clock rate to the maximum clock rate, resource  utilization measures the fraction of multipliers and adders utilized in the design relative to those available in DSP blocks, and compute efficiency reflects the fraction of useful MAC cycles achieved.

Optimizing clock rates and improving resource utilization often represent competing strategies for achieving high-performance FPGA accelerators. Note that logic resources are typically excluded from peak performance calculations, as their usage complicates computation, requires significant resources for auxiliary functions, and negatively impacts the working clock rate \cite{parker2014understanding}.

Overall efficiency provides a unified view of hardware performance, helping identify imbalances across components. By incorporating previously discussed metrics, this evaluation framework facilitates a quantitative comparison of existing research and highlights emerging trends in CNN hardware accelerators.

\subsubsection*{\textbf{Energy Efficiency}}

Energy efficiency \cite{liu2021toward, fan2021high} measures how efficiently the CNN hardware accelerators perform computations relative to its power consumption. It is typically expressed as the amount of work completed per watt of power consumed. The formula for energy efficiency is:
\begin{equation}
\text{Energy EFF.} = \frac{\text{Throughput}}{\text{Power}}
\end{equation}

As computational demands grow, especially in resource-intensive tasks like large-scale data processing and CNN model training, managing power consumption has become increasingly critical. Optimizing hardware accelerators to minimize power consumption while maintaining high throughput is a key challenge in the field. Improving energy efficiency not only reduces operating costs but also addresses thermal management issues, ensuring better overall system stability and sustainability.

\subsection{Discussion on Evaluation Framework}

This section introduces various evaluation metrics for CNN hardware accelerators, enabling a quantitative comparison of performance across different designs in subsequent sections. Additionally, the relationships between these metrics are examined, providing valuable insights for the design of FPGA-based CNN hardware accelerators.

A significant observation from the literature review is that some researchers evaluate hardware performance using only a limited subset of metrics. To address this, Section \ref{sec:hardware}  compare hardware designs using the comprehensive metrics of computational efficiency and overall efficiency introduced in this section. This approach aims to provide readers with a clearer understanding of the trade-offs and performance characteristics of different designs, offering practical guidance for future research and development in CNN hardware accelerators.

\section{CNN Acceleration Methods}
\label{sec:methods}

CNN models that achieve high accuracy tend to be large and resource-intensive. However, the larger the model, the more memory and storage it requires, making deployment on resource-constrained devices a significant challenge. Furthermore, larger models typically result in higher inference times and increased energy consumption, which limits their practicality for real-world applications. While these models perform well in controlled environments, they are not always feasible for deployment on edge devices.
The solution lies in reducing the size of the model through compression techniques or optimizing the computations involved in CNNs.

In this section, we explore existing methods for accelerating CNNs. We offer both qualitative and quantitative analyses of each approach, helping readers select the techniques that best suit their specific system requirements. The common methods for CNN acceleration can be grouped into the following categories:

\begin{enumerate}

\item \textbf{Pruning and Quantization}: This approach reduces the computational complexity and memory demands of a model by either decreasing the number of model parameters or reducing the bit-width of data. Techniques such as sparsity, pruning, and quantization are commonly employed in this category.

\item \textbf{Model Structure Compression}: This method optimizes the model's structure to enhance performance. It includes techniques like lightweight networks, knowledge distillation, layer fusion, and the use of smaller convolution kernels.

\item \textbf{Computation Reduction}: This approach replaces complex operations in the original network with more efficient computational techniques, such as FFT-based convolutions, fully spectral methods, and Winograd convolutions.

\end{enumerate}

\subsection{Pruning and Quantization}

CNN models often contain a large number of parameters, which are adjusted during training to minimize the loss function. These parameters help the network capture intricate data patterns by connecting multiple layers of neurons and applying non-linear activation functions. While this enables the model to achieve high accuracy, it also introduces significant computational costs. Therefore, controlling the quantity and quality of parameters is crucial to balancing performance and computation workloads.

\subsubsection{\textbf{Pruning}}

Pruning reduces the size and complexity of CNN models by eliminating unimportant parameters or structures. Similar to "pruning branches" in biology, this technique removes redundant components to make the model more efficient. The process typically involves training a full-scale deep neural network, evaluating the importance of each parameter based on criteria such as weight magnitude or activation levels, and then removing unimportant weights, neurons, channels, or layers. The remaining network is usually fine-tuned to restore performance, with pruning and fine-tuning often repeated for further compression.

For example, Han \textit{et al.} \cite{han2015deep} combined pruning, quantization, and Huffman coding to efficiently compress deep neural networks, reducing memory usage and improving computational speed and energy efficiency. Liu \textit{et al.} \cite{liu2017learning} introduced L1 regularization for automatic identification and pruning of unimportant channels, applicable to modern CNN architectures with minimal computational overhead. Lin \textit{et al.} \cite{lin2020hrank} proposed a pruning criterion based on the rank of the output feature map to rank and prune filters efficiently. Yeom \textit{et al.} \cite{yeom2021pruning} used Layer-wise Relevance Propagation (LRP) to prune networks, calculating the importance of each unit through back-propagation for better compression and accuracy, especially without fine-tuning. Liu \textit{et al.} \cite{liu2020autocompress} proposed AutoCompress, an automatic structured pruning framework that optimizes the pruning rate for each layer. More recently, Tan \textit{et al.} \cite{tan2024data} developed the Moving-one-Sample-out (MoSo) method, which measures sample importance based on the change in empirical risk when a sample is excluded, reducing computational overhead while maintaining high performance, even at high pruning ratios.

\subsubsection{\textbf{Quantization}}

Quantization is a technique used to reduce the computational overhead and memory demands of neural networks while preserving accuracy. By converting model weights and activation values from floating-point numbers (typically 32-bit) to lower-precision integers (e.g., 8-bit or 16-bit), quantization reduces memory usage and accelerates inference speed. This technique is particularly valuable for deploying models on edge devices like FPGAs, where integer representations are more efficient than floating-point operations.

For FPGA-based accelerations, fixed-point quantization is commonly used. Nguyen \textit{et al.} \cite{nguyen2019high} proposed a high-throughput, low-power FPGA implementation of the YOLO CNN for object detection, using binary weights and flexible low-bit activations, reducing model size and hardware costs. Kim \textit{et al.} \cite{kim2021zero} developed a fixed-point quantization method for YOLO, adjusting weight distributions by subtracting the mean value before quantization and applying iterative retraining to minimize accuracy loss. 
For hardware implementation, Jain \textit{et al.} \cite{jain2020trained} introduced a uniform symmetric quantizer with power-of-two scaling factors to achieve better optimization between range and precision.
Jin \textit{et al.} \cite{jin2022f8net} proposed F8Net, a fixed-point 8-bit multiplication network quantization method that achieves superior performance and efficiency compared to existing methods. Lin \textit{et al.} \cite{lin2016fixed} introduced a fixed-point quantization method for deep convolutional networks (DCNs) based on optimizing the signal-to-quantization-noise ratio (SQNR) to determine the optimal bit width for each layer.

To maximize acceleration, some researchers compress networks to even lower bit widths. Li \textit{et al.} \cite{li2019fully} introduced FQN (Fully Quantized Network), which uses fully quantized training at low bit widths, ensuring equal distances between adjacent quantization points. Cao \textit{et al.} \cite{cao2019seernet} developed SeerNet, which uses highly quantized (e.g., 4-bit or 1-bit) versions of CNN networks to predict binary sparse masks for output feature maps, guiding full-precision convolutions to exploit sparsity and accelerate inference. Similarly, Rastegar \textit{et al.} \cite{rastegari2016xnor} proposed XNOR-Net, where both filters and inputs to convolutional layers are binary, achieving 58$\times$ faster convolutions and 32$\times$ memory savings.

\subsection{Model Structure Compression}

The complexity of neural network architectures stems from the design and interaction of key components such as depth, width, and branching structures. Network depth refers to the number of layers in a model, and deeper networks can capture more abstract, complex features. However, very deep models can suffer from gradient vanishing or exploding problems, making training more challenging. Network width is the number of neurons or channels per layer, which influences the model's ability to process and capture input data. While wider networks can increase the capacity to learn, they can also introduce parameter redundancy and higher memory usage. Additionally, branching architectures where a model has multiple branches at different stages can improve performance by enabling the model to capture multi-scale features. However, these designs also increase computational complexity and memory demands.

In general, more complex network architectures can yield better performance but at the cost of higher computational overhead and memory requirements. When deploying these models on resource-constrained edge devices, there is a need to balance complexity with efficiency. Model structure compression aims to reduce the computational burden and memory usage by optimizing the network's design or eliminating redundant components. These techniques focus on the architectural level of the network and are essential for creating more efficient, deployable models.

\subsubsection{\textbf{Lightweight Networks}}
Lightweight networks are neural network models designed to operate efficiently in resource-constrained environments, aiming to reduce model size, computational complexity, and memory usage while preserving high performance.

Szegedy \textit{et al.} \cite{szegedy2015going} introduced GoogleNet, which featured the ``Inception module"—a multi-scale convolutional layer design that employs 1$\times$1, 3$\times$3, and 5$\times$5 convolutional kernels in parallel. This approach captures multi-scale features while keeping computational cost low. By repeating the Inception module across the network and combining it with pooling layers, they successfully increased both the depth and width of the model without significantly increasing the computational load.
Similarly, Iandola \textit{et al.} \cite{iandola2016squeezenet} proposed SqueezeNet, which uses a novel "Fire module" to build a lightweight model. The Fire module consists of a squeeze layer (using 1$\times$1 convolutions to reduce feature map dimensions) and an expand layer (combining 1$\times$1 and 3$\times$3 convolutions), making it efficient and compact.

The MobileNet series has also gained significant attention for its lightweight design. Howard \textit{et al.} \cite{howard2017mobilenets} introduced MobileNet v1, which uses depthwise separable convolutions, splitting traditional convolution into two parts: Depthwise Convolution and Pointwise Convolution. This reduces the number of parameters and computational complexity, while allowing flexibility in latency-accuracy trade-offs through two simple hyperparameters. MobileNet v2 \cite{sandler2018mobilenetv2} further optimized the network by introducing linear bottlenecks and inverted residuals, enabling deeper yet smaller and faster networks. MobileNet v3 \cite{howard2019searching} refined MobileNet v2 by removing certain layers, reducing computational overhead without sacrificing accuracy.

Han \textit{et al.} \cite{han2020ghostnet} proposed GhostNet, which uses the Ghost module for model compression. This technique splits a convolutional layer into two parts: one performing standard convolution and the other generating additional feature maps through simple linear operations, significantly reducing both computation and parameter count while maintaining accuracy.
Chen \textit{et al.} \cite{chen2023run} introduced Partial Convolution (PConv), which effectively extracts spatial features by reducing redundant computations and memory accesses. Based on PConv, FasterNet was also developed in \cite{chen2023run}, a new family of networks that achieves higher running speeds on various devices without sacrificing accuracy in visual tasks.

\subsubsection{\textbf{Knowledge Distillation}}

Knowledge Distillation (KD) is a model compression technique that enhances the efficiency of models without significant loss in performance. In this approach, a large, pre-trained, complex model (the teacher) transfers its knowledge to a simpler, smaller model (the student). The goal is to enable the student model to achieve performance close to that of the teacher model despite having fewer parameters. KD has been widely applied in areas such as mobile deployment, edge computing, real-time applications, resource conservation, and fields like medical imaging, natural language processing, and autonomous systems.

The concept of KD was first introduced by Hinton~\textit{et al.} \cite{hinton2015distilling}, who demonstrated how a teacher model could transfer knowledge to a smaller model by using both hard targets (the original labels) and soft targets (the teacher's outputs). This combination improves the accuracy of the student model. However, when there is a large size discrepancy between the teacher and student, the student model may struggle to learn effectively. To address this, Son \textit{et al.} \cite{son2021densely} proposed the use of multiple progressively smaller Teacher Assistants (TAs) to guide the student, making the knowledge transfer smoother. They also introduced the Random DGKD method, which randomly drops knowledge connections to avoid overfitting.

Hao \textit{et al.} \cite{hao2021self} advanced KD with Self-Mutual Knowledge Distillation (SMKD), where the model distills knowledge from itself, and student networks learn from each other in a mutual distillation process. This approach improves generalization and enhances both the visual and contextual capabilities of the network. Later, Sun \textit{et al.} \cite{sun2024logit} proposed a method where the temperature (scaling factor) of both the teacher and student models, as well as the sample logits, is varied. They introduced a Z-score preprocessing method that standardizes logits, enabling the student model to better learn the relationships from the teacher model.

\subsubsection{\textbf{Layer Fusion}}

In CNNs, layers such as convolution, batch normalization, activation, and pooling are typically executed in sequence. However, researchers have shown that reordering or merging certain layers can significantly reduce data transfer times and computational complexity when executing the CNN models in hardware.
Alwani \textit{et al.} \cite{alwani2016fused} introduced a pyramid-shaped multi-layer sliding window that processes input feature maps, allowing multiple layer results to be computed in advance. Abtahi \textit{et al.} \cite{abtahi2018accelerating} proposed integrating Batch Normalization (BN) layer parameters directly into the preceding convolutional layer during inference, as BN is a linear operation with fixed parameters. This integration reduces the number of operations. Syafeeza \textit{et al.} \cite{syafeeza2015convolutional} demonstrated that performing max pooling before ReLU activation helps reduce the computational workload on the ReLU layer. Liu \textit{et al.} \cite{liu2021accelerating} applied layer fusion to the Fast Fourier Transform (FFT) approach, combining pooling and convolutional layers to eliminate unnecessary operations and improve efficiency.

\subsection{Computation Reduction}
Computation reduction involves simplifying complex operations in neural networks, typically leveraging mathematical theorems. Techniques like the Fast Fourier Transform (FFT) and the Winograd algorithm can significantly reduce the computational complexity of convolutional layers. Further advancements, such as fully spectral CNNs, boost performance by transforming each layer into the frequency domain.

\subsubsection{\textbf{FFT Approach}}
The FFT approach accelerates convolution operations by using the convolution theorem, which states that spatial domain convolutions are equivalent to point-wise multiplications in the frequency domain. By transforming both the input data and weights into the frequency domain with FFT, performing element-wise multiplication, and then converting the result back to the spatial domain through Inverse FFT (IFFT), convolution operations can be executed more efficiently.

Mathieu \textit{et al.} \cite{mathieu2013fast} applied FFT to speed up both training and inference in convolutional neural networks, achieving notable improvements by reusing transformed feature maps. Rippe \textit{et al.} \cite{rippel2015spectral} introduced frequency domain pooling, observing that low-frequency components carry most of the critical information. By truncating feature representations in the frequency domain rather than using traditional spatial pooling, they preserved more information and allowed for flexible adjustments in the output size. Zeng \textit{et al.} \cite{zeng2018framework} implemented FFT on FPGA and developed a frequency domain loop tiling technique called Overlap-and-Add (OaA) to enhance throughput through better data reuse. Abtahi \textit{et al.} \cite{abtahi2018accelerating} compared three methods on embedded hardware: Direct Convolution (Direct-Conv), FFT-based Convolution (FFT-Conv), and Overlap-and-Add based FFT Convolution (FFT-OVA-Conv), providing a detailed comparison of computational complexity and memory requirements.

While traditional FFT methods reduce computational complexity, they have a significant drawback: the lack of frequency domain activation functions. This limitation necessitates repeated FFT-IFFT transformations, diminishing the acceleration benefits. To address this, fully spectral CNNs were introduced.
Liu \textit{et al.} \cite{liu2022design} overcame this limitation by constructing activation functions in the frequency domain, enabling all CNN layers to operate in the frequency domain and eliminating the need for repeated FFT-IFFT transformations. Similarly, Ayat \textit{et al.} \cite{ayat2019spectral} utilized the convolution theorem to create a frequency-domain SReLU activation function, further enhancing computation through layer fusion techniques. Watanabe \textit{et al.} \cite{watanabe2021image} explored the ReLU operation in the frequency domain and proposed the 2SReLU (Second Harmonics Superposition ReLU) activation function, performing all CNN operations in the frequency domain.

\subsubsection{\textbf{Winograd}}
While FFT is effective for large convolution kernel operations, CNNs increasingly use smaller kernels, where the Winograd algorithm excels. The Winograd algorithm minimizes the number of multiplication operations in convolution by substituting multiplications with additions, thus reducing computational complexity.

Although originally proposed by Shmuel Winograd in 1980, the algorithm was first applied to accelerate convolution operations in neural networks by Lavin \textit{et al.} \cite{lavin2016fast}. They divided large input data into smaller tiles and performed low-complexity convolutions on each tile, significantly reducing computational load. Their experiments demonstrated the algorithm's efficiency, particularly for smaller convolution kernels. Subsequently, Liang \textit{et al.} \cite{liang2019evaluating} developed a hardware framework to implement the Winograd algorithm on FPGAs, utilizing a line-buffer structure for efficient feature map reuse across different blocks, and pipelining Winograd Processing Element (PE) units in parallel. Yepez \textit{et al.} \cite{yepez2020stride} introduced a novel Winograd-based method that supports strides of 2 and is adaptable across one, two, or three dimensions.

\subsection{Discussion and Comparison}

\begin{table*}[t]

\centering

\caption{Comparison of CNN Acceleration Methods}

    \renewcommand\arraystretch{2.6}
    \resizebox{0.95\textwidth}{!}
    {
    \begin{tabular}{|m{3cm}<{\centering}|m{1cm}<{\centering}|m{9cm}|m{7cm}|}
    \hline
        \textbf {Type} & \textbf {Ref.} & \multicolumn{1}{|c|}{\textbf{Methods}} & \multicolumn{1}{|c|}{\textbf {Results}}  
        \\ \hline
         \multirow{4}{*}{\textbf{Pruning}} &\cite{han2015deep} & 
         Combined pruning, quantization, and huffman coding. & \makecell[l]{
         $\bullet$ AlexNet parameters reduced: 97.12$\%$ \\  $\bullet$ VGG-16 parameters reduced: 97.95$\%$ }
    \\ \cline{2-4}  
         & \cite{lin2020hrank} & Used the rank of the output feature map to determine the ranking of filters. & \makecell[l]{
         $\bullet$ ResNet-110 parameters reduced: 59.2$\%$ \\  $\bullet$ ResNet-50 parameters reduced: 36.7$\%$}  
    \\ \cline{2-4}  
         & \cite{liu2017learning} & Used L1 regularization to the batch normalization layers. & $\bullet$ VGGNet parameters reduced: 95$\%$
    \\ \cline{2-4} 
         & \cite{liu2020autocompress} & Proposed an automatic structured pruning framework. & $\bullet$ VGG-16 and ResNet-18, parameters reduced: 99.16$\%$.
    \\ \hline   
        \multirow{4}{*}{\textbf{Quantization}} & \cite{kim2021zero} &  Centered the weight distribution to zero. & $\bullet$ YOLOv3 and YOLOv4, parameters reduced: 80$\%$.
    \\ \cline{2-4}  
         & \cite{lin2016fixed} & Used SQNR to find the optimal bit width for each layer. &  $\bullet$ DCNs parameters reduced: 20$\%$ 
    \\ \cline{2-4}  
         & \cite{rastegari2016xnor} & Both the filters and the input to convolutional layers are binary. & $\bullet$ ResNet18 parameters reduced:: 98.5$\%$  
    \\ \cline{2-4}   
         & \cite{cao2019seernet} & Proposed a binary sparsity by inference on the original network. & $\bullet$ ResNet18 speedup: 2.45$\times$
    \\ \hline   
        \multirow{5}{*}{\textbf{Lightweight Networks}} & \cite{szegedy2015going} & Introduced the "Inception module". & \makecell[l]{
         $\bullet$  Top-1 accuracy  on ImageNet: 43.9$\%$ \\  $\bullet$  Parameters: 5 million}   
    \\ \cline{2-4}   
         & \cite{iandola2016squeezenet} & Introduced the fire module. & \makecell[l]{
         $\bullet$  Top-1 accuracy  on ImageNet: 57.5$\%$ \\  $\bullet$  
          Parameters: 4.8 million}
    \\ \cline{2-4}  
         & \cite{howard2017mobilenets} & Introduced depthwise separable convolutions. & \makecell[l]{
         $\bullet$  Top-1 accuracy  on ImageNet: 70.6$\%$ \\  $\bullet$  Parameters: 4.2 million} 
    \\ \cline{2-4}  
         & \cite{han2020ghostnet} & Introduced the ghost module for model compression. & \makecell[l]{
         $\bullet$  Top-1 accuracy  on ImageNet: 73.9$\%$ \\  $\bullet$ Parameters: 5.2 million}
    \\ \cline{2-4}   
         & \cite{chen2023run} &  Proposed a new partial convolution (PConv). & \makecell[l]{
         $\bullet$  Top-1 accuracy  on ImageNet: 76.2$\%$ \\  $\bullet$ Parameters: 7.6 million} 
     \\ \hline  
        \multirow{3}{*}{\textbf{FFT}} & \cite{zeng2018framework} & Developed overlap-and-add technique. & \makecell[l]{
         $\bullet$ AlexNet speedup: 10.6$\times$ \\  $\bullet$ VGG16 speedup: 7.4$\times$ }
    \\ \cline{2-4}  
         & \cite{liu2022design} & Introduced spectral activation functions. & \makecell[l]{
         $\bullet$ Speedup compared to spatial methods: $4 \sim 6.6\times$ \\ $\bullet$ Speedup compared totraditional FFT: $3 \sim 4.4\times$}
    \\ \cline{2-4}  
         & \cite{ayat2019spectral} & Introduced SReLU activation function. & $\bullet$ Speedup compared to spatial methods: 3.45$\times$ 
    \\ \hline  
        \multirow{3}{*}{\textbf{Winograd}} & \cite{lavin2016fast} & Used winograd to accelerate convolution operations. & $\bullet$ Speedup: $1.63 \sim 7.42\times$ 
    \\ \cline{2-4}
        & \cite{liang2019evaluating} & Developed a hardware
        framework for w
        inograd algorithm. & $\bullet$ AlexNet speedup: 11.8$\times$ 
    \\ \cline{2-4}
        & \cite{yepez2020stride} & Introduced the winograd algorithm to more dimensions. & $\bullet$ Speedup: $1.44 \sim 2.42\times$     
     \\ \hline  
    \end{tabular}
    }    

\vspace{-2ex}
    \label{tab:2}
\end{table*}

We have reviewed the current methods for accelerating CNNs as described above. Here, we also provide a comprehensive summary of their key research efforts, as shown in Table~\ref{tab:2}. This summary highlights each method's core contributions and results, aiming to offer an overview of their respective acceleration effects. However, each method has its own set of advantages and limitations, making them more or less suitable depending on the specific scenario.

For example, both pruning and quantization can help reduce computational overhead. However, pruning is a complex process that often depends on the characteristics of the task and hardware architecture. Moreover, it may require careful fine-tuning. Quantization, while effective at reducing model size and computation, may require additional calibration steps to maintain performance.

Knowledge distillation allows smaller models to learn from larger, more complex models, leveraging their output distributions or intermediate features. While this can result in compact models with good performance, the distillation process itself is relatively complex and heavily reliant on the teacher model and the specific task at hand. Lightweight models, on the other hand, are designed to be compact and efficient, often using specialized architectures to reduce memory and computation. However, they may not match the performance of larger, general-purpose models in terms of accuracy.

The FFT approach can significantly reduce the computational complexity of convolution layers by leveraging frequency domain operations. However, it requires repeated FFT and IFFT transformations, which can hinder its effectiveness in hardware implementations. Fully spectral CNNs, which aim to overcome this limitation, have been optimized for hardware, though they may not be ideal for scenarios with large inputs. The Winograd algorithm reduces convolution complexity by minimizing the number of multiplications required, improving computational efficiency. However, it introduces additional intermediate transformation matrices and cache requirements, leading to higher memory usage.

To determine the most optimal acceleration method for a given scenario, we conduct both qualitative and quantitative analyses of these methods, as shown in Table \ref{tab:1}. 
The speedups for each method are estimated based on the results presented in the literature.
This table provides valuable insights into the strengths and weaknesses of each approach, helping us identify the most suitable method for maximizing acceleration in various use cases.

\begin{table*}[t]
  \centering

\caption{Summary and Quantitative Analysis of the Reviewed Methods}

    \renewcommand\arraystretch{1.0}
    \resizebox{0.95\textwidth}{!}
    {
    \begin{tabular}{|c|c|c|c|c|c|}
        \hline
        Methods &  Cheaper Arithmetic Operations & Memory Reduction & Parameters Reduction & Hardware Friendly & Speedup 
        \\  \hline
        
        Pruning  & $\times$ & $\checkmark$ & $\checkmark$ & $\checkmark$ & $2 \sim 10 \times$ \\
        Quantization & $\checkmark$ & $\checkmark$ & $\times$  & $\checkmark$ & $2 \sim 50 \times$ \\
        Lightweight Networks & $\times$ & $\checkmark$ & $\checkmark$ & $\checkmark$ & $ 8 \sim 9 \times$ \\
        Knowledge Distillation & $\times$ & $\checkmark$ & $\checkmark$ & $\checkmark$ & $ 1.5 \sim 5 \times$ \\
        Layer Fusion & $\checkmark$ & $\checkmark$ & $\times$ & $\checkmark$ & $\geq 2 \times$\\
        FFT & $\checkmark$ & $\times$ & $\times$ & $\times$ & $5 \sim 10 \times$\\
        Winograd & $\checkmark$ & $\times$ & $\times$ & $\checkmark$ & $3 \sim 5 \times$\\
        \hline
    \end{tabular}
    }

\label{tab:1}
\end{table*}
\section{Hardware Approaches}
\label{sec:hardware}

This section introduces the key techniques commonly used for CNN hardware acceleration, followed by an overview of optimization strategies that build upon these techniques to enhance performance.

\subsection{Data Blocking and Parallel Computing}

\subsubsection{\textbf{Loop Tiling}}

Loop tiling, also known as data blocking, is a critical optimization technique in hardware acceleration \cite{yang2023loop, huang2022efficient}, especially for CNN models that involve large-scale data processing. Its main objective is to minimize memory access overhead and maximize hardware resource utilization by enhancing cache efficiency.

In convolutional nested loops, large input feature maps are typically processed element by element. While this method is straightforward, it often leads to frequent cache misses, as large portions of data do not remain in the processor’s cache. This results in excessive memory accesses, which degrade overall performance. Loop tiling addresses this issue by dividing large loops into smaller, more manageable tiles, as shown in Code \ref{alg:cnn2}, where $T_i$ and $T_j$ denote tiling coefficients for different loop levels. By processing only a subset of data at a time, loop tiling ensures that each tile remains in the cache, significantly improving data locality and reducing memory bandwidth usage.
The technique introduces additional outer loops that partition the dataset into smaller chunks. Each chunk is processed independently, allowing it to be efficiently stored in cache and reducing memory access overhead. This segmentation enhances execution speed by minimizing data movement between main memory and processing units.

In hardware-accelerated environments, loop tiling is often combined with parallelism and pipelining to further optimize performance. In parallel computing architectures such as GPUs, multiple processing units can operate on different tiles concurrently, enabling efficient parallel execution. In FPGAs and custom accelerators, loop tiling can be integrated with hardware-specific architectural features to mitigate memory bottlenecks and enhance overall data throughput.

\begin{algorithm}[t]
\caption{Tiling Convolutional Layer Algorithm}
\label{alg:cnn2}
\begin{algorithmic}[1]
 \STATE for {($f=0$ ; $f<F$ ; $f++$)}  \quad \quad \quad \quad 
 \STATE \quad  for {($c=0$ ; $c<C$ ; $c++$)} \quad \quad \quad 
 \STATE \quad \quad  for {($h=0$ ; $h<H_{o}$ ; $h++$)} \quad  
 \STATE \quad\quad\quad   for {($w=0$ ; $w<W_{o}$ ; $w++$)} 
 \STATE \quad\quad\quad\quad  for {($ii=0$ ; $ii<K$ ; $i+T_i$)} 
 \STATE \quad\quad\quad\quad  for {($jj=0$ ; $jj<K$ ; $j+T_j$)} 
 \STATE \quad\quad\quad\quad  for{($i=ii$ ; $i<ii+Ti$ ; $i++$)}
 \STATE \quad\quad\quad\quad  for{($j=j$ ;  $j<jj+Tj$ ; $j++$)}
\STATE 
            $$O[f][h][w]+=W[f][c][i][j]*I[c][h*S+i][w*S+j]$$
\end{algorithmic}
\end{algorithm}

\subsubsection{\textbf{Loop Unrolling}}

Loop unrolling is a widely used optimization technique aimed at improving the execution efficiency of loops in CNN hardware accelerators, particularly on FPGAs. The key idea is to merge multiple iterations of a loop into a single, larger operation, reducing loop control overhead and increasing parallelism to accelerate execution.

In CNN computations, convolution operations often involve deeply nested loops, leading to a high number of iterations when processing large-scale input data. Loop unrolling addresses this inefficiency by transforming a single iteration into multiple parallel operations. By increasing the workload of each iteration, it reduces the need for control logic, such as loop counters and branching decisions, thereby improving instruction execution efficiency and maximizing hardware parallelism.
For instance, in convolution operations, a standard loop structure processes input data elements sequentially. With loop unrolling, multiple elements can be processed in a single iteration, or different convolution tasks can be executed simultaneously, significantly boosting computational throughput.

However, the degree of unrolling must be carefully tuned to match the available hardware resources, such as logic units and memory bandwidth. Excessive unrolling can lead to resource congestion, negatively affecting overall performance. Therefore, in CNN hardware acceleration, loop unrolling must be strategically applied to strike a balance between performance gains and efficient hardware utilization.

\begin{figure}[t]
     \centering
       \includegraphics[width=0.45\textwidth]{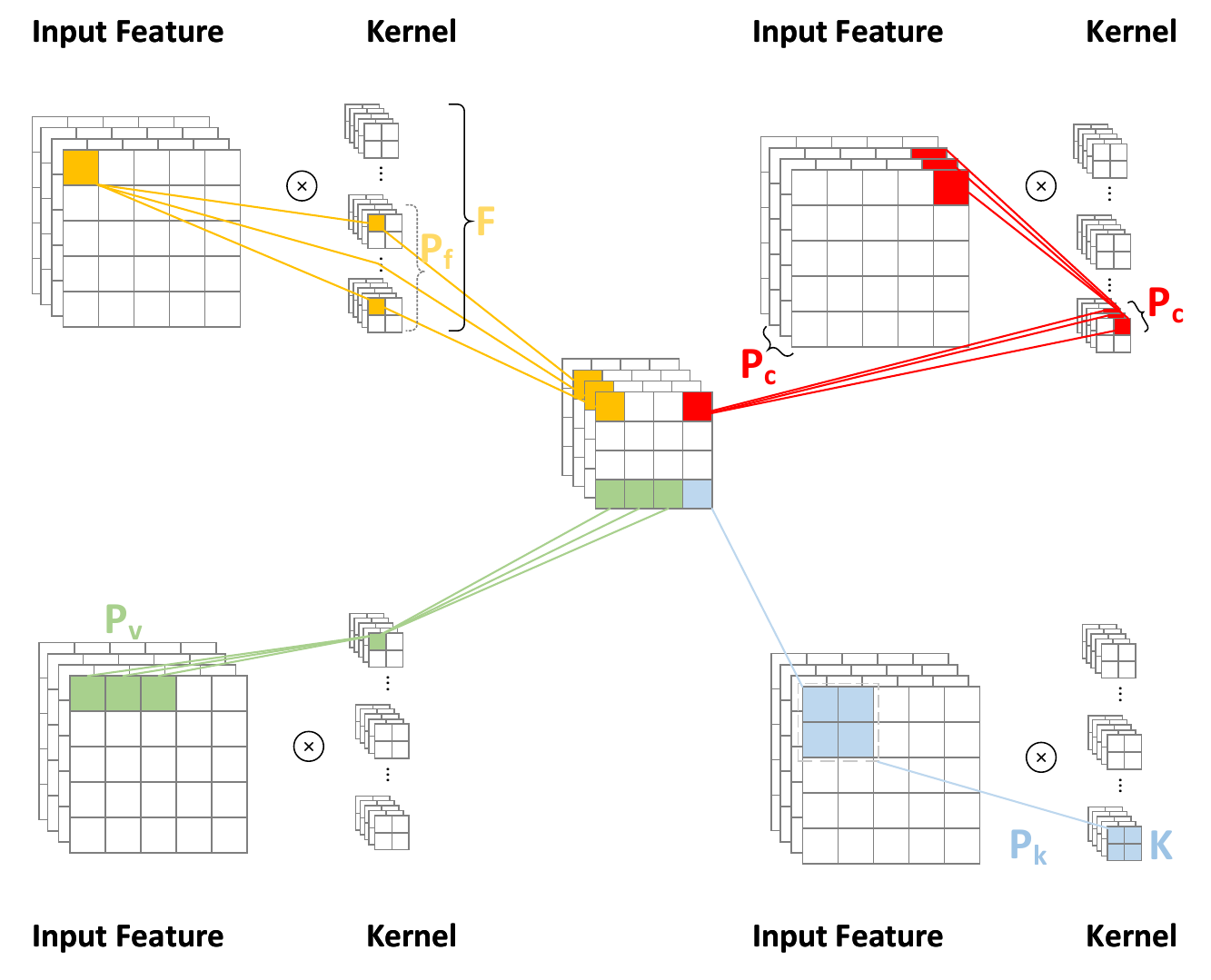} 
       \vspace{-1ex} 
    \caption{An illustration depicting the four levels of parallelism in convolution computation across the filter ($P_f$), channel ($P_c$), pixel ($P_v$), and kernel ($P_k$) dimensions.}
    \vspace{-2ex}
    \label{fig:Para}
\end{figure}

\subsubsection{\textbf{Parallel Computing}}

Parallel computing is a fundamental technique in CNN hardware acceleration, particularly for computationally intensive tasks like convolution operations. Convolutions involve numerous multiplications and additions, which can be executed concurrently across multiple processing units, significantly boosting computational speed.

In hardware acceleration platforms such as FPGAs and GPUs, convolution operations are naturally suited for parallel execution. By distributing data and tasks across multiple computing units, processing throughput is greatly enhanced. For example, FPGAs can be configured with numerous parallel computing units to process multiple input data points and convolution kernel elements simultaneously, alleviating the bottleneck of sequential execution. In such architectures, multiplication and summation of input data and convolution kernel elements occur in parallel, fully utilizing the available hardware resources.
Additionally, the reconfigurability of FPGAs allows for the customization of parallel computing architectures tailored to specific CNN models. Techniques such as pipelined data flow processing and hierarchical parallel task scheduling can further enhance CNN inference speed. Moreover, loop unrolling, which increases computational workload per iteration while reducing loop control overhead, creates more opportunities for parallel execution, further improving efficiency.

Fig. \ref{fig:Para} illustrates several parallel computing strategies commonly used to accelerate convolution operations. Convolution layers consist of multiple nested loops that slide over the kernel and feature maps, offering a vast design space for parallel computing strategies and data caching optimization. These loop transformations have been widely explored in previous research. Expanding different loops directly impacts parallel computing methods, memory access patterns, and convolution engine architectures. To simplify the discussion, we define parallelism in different dimensions as follows:

\textbf{(1) Filter Parallelism ($P_f$):} In CNNs, each input feature map is convolved with multiple filters to generate corresponding output feature maps. Traditionally, this process is performed sequentially, which underutilizes hardware resources. With filter parallelism, multiple filters operate simultaneously on the same input feature map, generating multiple output channels in parallel. This approach enables efficient reuse of feature data and significantly enhances computational throughput.

\textbf{(2) Channel Parallelism ($P_c$):} CNNs typically process multi-channel input feature maps, such as RGB images (three channels) or deeper feature maps generated by previous convolution layers. Conventionally, each convolution kernel computes results for each channel independently before summing them to produce the final output. Channel parallelism accelerates this process by simultaneously computing convolutions across all input channels, reducing computational dependencies and improving efficiency. This method also facilitates effective reuse of weight data.

\textbf{(3) Pixel Parallelism ($P_v$):} In convolution operations, kernels slide across input feature maps to compute convolution results at different spatial positions. Since these computations at different positions are largely independent, they can be processed in parallel. Pixel parallelism exploits this independence by computing multiple spatial locations simultaneously. While this approach simplifies data storage, it requires multiple levels of buffering for data shifting in on-chip cache design due to the sliding window nature of convolutions.

\textbf{(4) Kernel Parallelism ($P_k$):} Kernel parallelism extends pixel parallelism to the convolution kernel dimension. It processes multiple weight values for a single feature pixel in parallel. However, this method is often constrained by kernel size and presents challenges in data storage design, making it less flexible for large-scale CNN architectures.

\begin{table*}[t]

\centering

\caption{Comparison of Hardware Performance of Existing CNN Accelerators using Different Parallelism}

    \renewcommand\arraystretch{2.0}
    
    \resizebox{0.95\textwidth}{!}
    {
    
    \begin{tabular}{|c|m{1.5cm}<{\centering}|c|m{1.5cm}<{\centering}|m{1.5cm}<{\centering}|m{1.5cm}<{\centering}|m{1.5cm}<{\centering}|m{1.5cm}<{\centering}|m{1.5cm}<{\centering}|m{1.5cm}<{\centering}|}
    \hline
        \textbf {Ref.} &  \textbf {Employed Parallelism}  & \textbf {Latency (ms)} & \textbf {Throughput (Gop/s)}  &  \textbf {Compute EFF.}  & \textbf {Resource EFF. (Gop/s/DSP)}  & \textbf {Clock EFF.} &  \textbf {Overall  EFF.} & \textbf {Power (W)} & \textbf {Energy EFF. (Gop/s/W)} 
        \\ \hline

        \cite{liu2018optimizing} & \multirow{2}{*}{\textbf{$P_k, P_v, P_f$}} & - & 107 & 41.7\% & 0.12 & 40\% & 16.6\% & 9.6 & 11.14
        \\ \cline{3-10}
        
        \cite{zhao2017optimizing}& & 3.7 & 391.6 & - & - & 40\% & - & 178 & 2.19 
        \\ \hline
               
        \cite{rahman2016efficient} & \multirow{2}{*}{\textbf{$P_v, P_f$}} & - & 147.82  & 17.14\% & 0.11 & 26.7\% & 4.24\% & - & -
        \\ \cline{3-10}

        \cite{ma2017end} & & 71.71 & 315.48 & 69.28\% & 0.21 & 33.3\% & 20.6\% & - & -
        \\ \hline 

        \cite{fan2021high} & \multirow{6}{*}{\textbf{$P_c, P_f$}} & 4.62 & 1667 & 92.05\% & 1.24 & 44\% & 36.1\% & 45 & 37 
        \\ \cline{3-10}

        \cite{liu2021toward} & & 5.07 & 1519 & 90\% & 1.0 & 44\% & 54\% & 19.1 & 79.5 
        \\ \cline{3-10}

        \cite{zhang2015optimizing} & & 21.61 & 61.62 & 69.1\% & 0.03 & 20\% & 11.04\% & 18.61 & 3.31 
        \\ \cline{3-10}

        \cite{ma2017optimizing} & & 47.97 & 645.25 & 68.58\% & 0.43 & 30\% & 20.57\% & - & -
        \\ \cline{3-10}

        \cite{yu2019opu} & & - & 1218 & 95\% & 0.76 & 40\% & 23.18\% & 17.7 & 68.81 
        \\ \cline{3-10}

        \cite{zhou2024design} & & - & 1240.8 & - & 0.32 & 28.6\% & - & 13.38 & 92.4 
        \\ \hline   
        
       \cite{wu2024efficient} & \textbf{$P_v, P_c, P_f$} & - & 2870 & 92.2\% & 0.79 & 38\% & 24.88\% & 39.5 & 72.7
       \\ \hline

        \cite{guo2017angel} & \multirow{4}{*}{\textbf{$P_k, P_c, P_f$}}& 364 & 84.3 & 80.25 \% & 0.17 &26.7\% & 18.58\% & 3.5 & 24
        \\ \cline{3-10}
        
        \cite{li2016high} & & 2.56 & 565.94 & 84.6\% & 0.26 & 31.2\% & 15.84\% & 30.2 & 18.74 
        \\ \cline{3-10}

        \cite{suda2016throughput} & & 20.1 & 136.5 & - & 0.56 & 24\% & - & 24.2 & 5.64
        \\ \cline{3-10}

        \cite{li2024high} & & 0.62 & 2423.63 & - & 0.89 & 48.2\% & - & 18.04 & 143.32 
        \\ \hline

        \cite{shah2018runtime} & \multirow{3}{*}{\textbf{Dynamic}} & - & 129.7 & 41.8\% & 0.13 & 30\% &4.63\% & 18 & 7.2 
        \\ \cline{3-10}

        \cite{khan2023towards} & & 178 & 684 & 49.47\% & 0.2 & 40\% & 19.39\% & 26 & 26.3 
        \\ \cline{3-10}

        \cite{dai2024dcp} & & 38.3 & 807 & 98.5\% & 0.79 & 40\% & 35.03\% & 6.3 & 128.1 
        \\ \hline
    \end{tabular}
    }

    \label{tab:performance}
\end{table*}

\subsection{CNN Hardware Accelerators based on Parallel Computing}

This section explores and analyzes the integration of data-level parallel computing strategies in the design of CNN hardware accelerators, as discussed in existing literature. Various parallelization techniques are employed in different combinations to optimize accelerator architectures \cite{dai2024dcp, fan2021high, guo2017angel, khan2023towards, li2016high, li2024high, liu2018optimizing, liu2021toward, ma2017end, ma2017optimizing, rahman2016efficient, shah2018runtime, suda2016throughput, wu2024efficient, yu2019opu, zhang2015optimizing, zhao2017optimizing, zhou2024design}.
This work presents a detailed comparison of various parallelization strategies, evaluating their computational efficiency and summarizing key performance metrics. The goal is to provide insights into the latest advancements in CNN hardware acceleration, enabling researchers, developers, and engineers to better understand the trade-offs among different approaches. By examining the characteristics and capabilities of diverse parallel strategies, readers can design optimized acceleration solutions tailored to specific tasks. A comprehensive summary of performance comparisons is provided in Table \ref{tab:performance}.

For example, Zhao \textit{et al.} \cite{zhao2017optimizing} and Liu \textit{et al.} \cite{liu2018optimizing} implemented a combination of kernel, pixel, and filter parallelism ($P_k$, $P_v$, $P_f$) to accelerate CNN computations. Liu \textit{et al.} achieved a processing performance of 107 Gop/s on a Zynq ZC7045 FPGA with 16-bit data quantization, reaching an efficiency of 0.12 Gop/s per DSP. While this approach delivered high computational throughput, it suffered from inefficient resource utilization, ultimately limiting overall accelerator efficiency.

Similarly, Rahman \textit{et al.} \cite{rahman2016efficient} and Ma \textit{et al.} \cite{ma2017end} adopted a hybrid strategy combining pixel parallelism ($P_v$) and filter parallelism ($P_f$). In their implementation of ResNet-50 on Intel FPGA Stratix V and Arria 10, Ma \textit{et al.} achieved a throughput of 315.5 Gop/s. This method maximized data reuse and improved computational efficiency. However, it introduced challenges in hardware resource utilization, particularly when convolution layer dimensions were not divisible by $P_v$, leading to underutilized computational units in certain CNN layers.


Unlike $P_k$ and $P_v$, which are constrained by kernel size and feature characteristics, the combined approach of $P_c$ and $P_f$ offers greater flexibility \cite{fan2021high, liu2021toward, zhang2015optimizing, ma2017optimizing, yu2019opu, zhou2024design}. For example, \cite{fan2021high} achieves energy efficiency $3.8 \sim 5.6\times$ higher than GPUs and up to $1.4 \sim2.2\times$ better resource efficiency than FPGA-based accelerators for 2D and 3D CNNs, with an overall efficiency of 50.47\%.

Some studies integrate pixel, channel, and filter parallelism ($P_v$, $P_c$, $P_f$) \cite{wu2024efficient}, achieving high throughput. For example, \cite{wu2024efficient} reported 2870 Gop/s peak performance for convolution operations. Channel parallelism enables large-scale parallelism more efficiently than pixel parallelism, improving throughput. Other works employ ($P_k$, $P_c$, $P_f$) \cite{guo2017angel, li2016high, suda2016throughput, li2024high}, extending parallelism to additional dimensions for better performance. However, $P_c$ and $P_f$ parallelism face constraints; for example, the first convolution layer typically processes RGB channels, but $P_c$ often scales exponentially by 2, leading to computational imbalances and reduced efficiency.

Despite various parallel strategies, architectures exhibit significant differences in compute efficiency due to fixed parallelism in each data dimension, resulting in resource under-utilization and limited adaptability. To address this, some works propose dynamic parallelism \cite{khan2023towards, dai2024dcp, shah2018runtime}. For example, DCP-CNN \cite{dai2024dcp} dynamically adjusts parallelism based on input size, kernel size, and network configuration, achieving over 800 Gop/s throughput and $72\% \sim 98\%$ compute efficiency on Intel Stratix 10 GX650 FPGA,  outperforming existing accelerators.
Dynamic parallelism optimizes FPGA-based CNN acceleration by tailoring parallel strategies to each layer, preventing load imbalance and enhancing efficiency. However, it introduces design complexity, requiring adaptive data storage and on-chip buffering to handle transitions between parallel strategies. Additionally, varying network parameters pose challenges in identifying optimal configurations.



\subsection{Other Optimizations}

\subsubsection{\textbf{Input Reshaping}}

Input reshaping is a key optimization in CNN hardware accelerators that enhances compute efficiency by reorganizing data storage and access patterns to better align with hardware architectures \cite{zhang2024design, li2022fpga, ting2018system, chen2021hardware}. It ensures efficient data alignment for convolution operations, facilitating optimized memory access and computation.

A common approach, Im2Col \cite{wang2021optimization}, converts feature maps into matrix form, enabling efficient matrix multiplications instead of direct convolution operations. Additionally, input reshaping can partition data into cache-friendly blocks, reducing memory accesses and improving processing efficiency. Format conversions, such as between $NCHW$ and $NHWC$, further optimize performance across different hardware architectures. For sparse inputs, compressed storage formats enhance sparse matrix computation efficiency.
While input reshaping improves memory utilization and supports parallelism, it may introduce computational and storage overhead. Effective integration with hardware design is essential to balance reshaping costs with performance gains.

\subsubsection{\textbf{DSP Optimization}}

CNN accelerators benefit from reduced precision due to their tolerance to small numerical errors, making low-precision computing a key strategy for improving performance. However, CNN operations rely on DSP units, which support limited precision levels, posing challenges for precision reduction. To address this, researchers have explored executing multiple low-precision MAC operations within a single DSP unit to enhance efficiency.

Lee \textit{et al.} \cite{lee2018double} introduced a 2-way SIMD MAC design for CNN convolutions, where SIMD multiplications share a common operand, enabling efficient dual MAC operations in a single DSP block. Similarly, Liu \textit{et al.} \cite{liu2021toward} proposed an INT8 optimization technique for Intel DSPs, utilizing an $18 \times 19$ multiplier to achieve a 1:2 DSP-to-INT8 MAC ratio. By partitioning INT8 computations into smaller multiplications, they optimized DSP usage while leveraging logic resources for additional operations.

\subsection{Discussion based on Our Evaluation Framework}

In this section, we provide a comparison of various works based on different parallel combinations, as shown in Table \ref{tab:performance}. A key observation from existing research is that many studies evaluate their designs using a limited set of metrics, such as throughput, power consumption, and latency. However, these metrics are not solely influenced by the design itself; they are also highly dependent on the FPGA device used, as performance can vary across different CNN models.
To address this, we emphasize the importance of compute efficiency and overall efficiency. These metrics, included in Table \ref{tab:performance} based on reported design parameters, offer a more comprehensive understanding of the evaluated works. Some entries are left blank because the results were not provided in the respective papers.

Compute efficiency is closely tied to the number of  MAC operations in the design and is relatively unaffected by FPGA device choice. We calculate and summarize compute efficiency from various hardware architectures, such as \cite{rahman2016efficient} (17.14\%) and \cite{dai2024dcp} (98.5\%). A key conclusion from our comparison is that parallelization strategy plays a significant role in determining compute efficiency. For example, \cite{rahman2016efficient} used a fixed parallelization strategy with constant parallelism degrees ($P_v$, $P_f$), which can lead to load imbalances and resource inefficiency. In contrast, \cite{dai2024dcp} dynamically adjusted the parallelization strategy based on the layer being executed, optimizing resource usage and achieving higher compute efficiency.

Given these findings, future research should focus on leveraging dynamic parallelism to maximize performance. However, implementing dynamic parallelism introduces design complexities, such as the need for flexible data storage schemes and on-chip buffers to adapt to different parallel strategies. Failure to properly manage these transitions may degrade data transfer efficiency and overall performance.

Overall efficiency builds upon compute efficiency but also considers resource utilization and clock efficiency. One key method to improve resource utilization is DSP optimization, which maximizes the use of limited DSP resources. For example, \cite{liu2021toward} demonstrated high overall efficiency largely due to effective DSP optimization. Improving clock efficiency requires careful attention to pipelining, timing optimization, and resource sharing.

\begin{figure*}[t]
     \centering
       \includegraphics[width=0.8\textwidth]{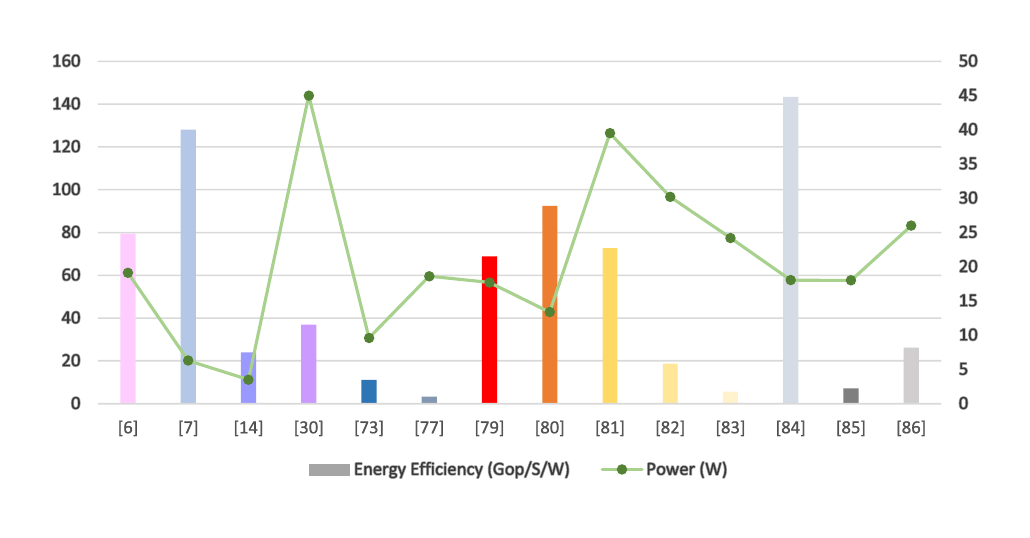} 
       \vspace{-6ex} 
    \caption{Energy efficiency comparison across CNN accelerator architectures. The left vertical axis represents energy efficiency, while the right vertical axis shows power consumption.}
    \vspace{-2ex}
    \label{fig:Energy Comparision}
\end{figure*}

We also compare the energy efficiency across designs, shown in Fig. \ref{fig:Energy Comparision}. Few designs achieve high energy efficiency with low power consumption; for most, energy efficiency is directly proportional to power consumption, meaning higher power usage leads to higher throughput. Thus, research efforts should focus on balancing low power consumption with high throughput.

\begin{figure*}[t]
     \centering
       \includegraphics[width=0.8\textwidth]{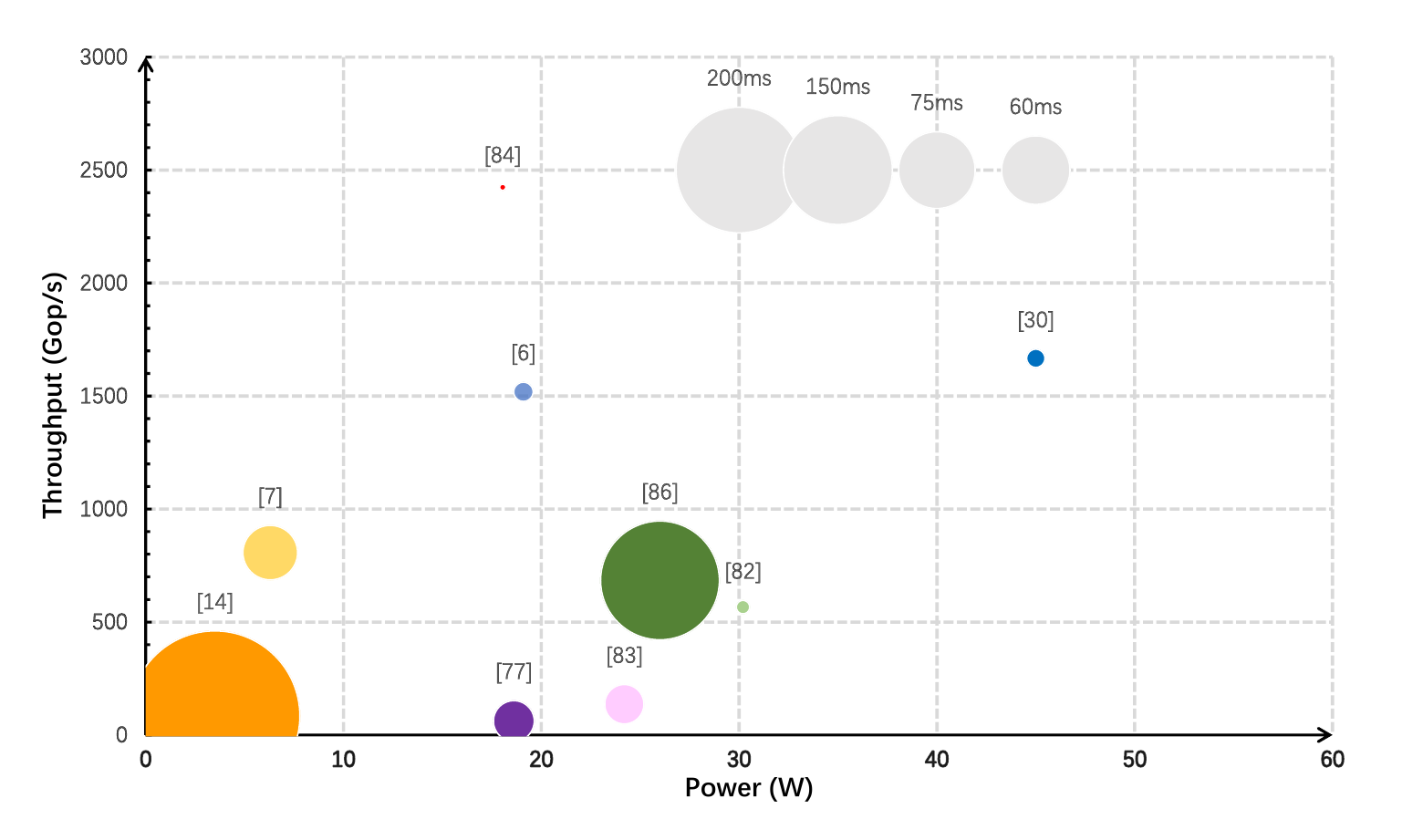} 
       \vspace{-2ex} 
    \caption{An illustrative comparison of the hardware performance of existing CNN accelerators in terms of power consumption, latency, and throughput. The size of the bubbles corresponds to latency, with larger bubbles indicating higher latency and smaller ones indicating lower latency. The horizontal axis represents power consumption, increasing from left to right, while the vertical axis represents throughput, increasing from bottom to top.}
    \vspace{-2ex}
    \label{fig:Pop1}
\end{figure*}

Additionally, we conduct a comprehensive comparison of latency, throughput, and power consumption for architectures based on different parallelism combinations. As shown in Fig.~\ref{fig:Pop1}, high throughput and low latency are generally associated with higher power consumption, while low-power designs tend to have lower throughput and higher latency. Future work should focus on optimizing architectures to enhance throughput while maintaining low power consumption.

\section{Algorithm-Hardware Co-Design }
\label{sec:co-design}

To meet the growing complexity and accuracy demands of neural architectures, the number of parameters in CNN models has increased significantly. While this boosts performance, it also raises computational and memory requirements, leading to higher time and energy consumption during training and inference. Edge computing platforms, such as mobile and embedded devices, face challenges due to limited processing power, storage, and energy efficiency, making it difficult to deploy these resource-intensive models, especially in real-time applications.

Optimizing only from the hardware or software perspective is not enough to address these challenges. Algorithm-hardware co-design integrates both aspects, tailoring algorithms to leverage hardware features while adapting hardware architectures to meet algorithmic needs. This approach maximizes performance, resource utilization, and energy efficiency by optimizing the synergy between software and hardware.

\subsection{CNN Acceleration Toolflows}

Toolflows represent a system-level approach to automating the mapping of CNNs onto FPGAs. These toolflows streamline the design process, ensuring rapid deployment and high energy efficiency.
As application needs and hardware capabilities evolve, CNN toolflows have adapted to drive continuous innovation in both algorithms and hardware. They encompass a wide range of technologies, from deep learning frameworks to hardware acceleration, automated optimization, and explainability analysis.
These tools enable users to generate customized CNN hardware implementations without requiring deep hardware expertise, making FPGA integration more accessible within the deep learning ecosystem. Table \ref{tab:hh} presents various CNN-to-FPGA toolflows that leverage the algorithm-hardware co-design framework to generate optimized FPGA accelerators tailored to specific CNN-FPGA pairs.

fpgaConvNet \cite{venieris2016fpgaconvnet} utilized specialized hardware blocks to efficiently map irregular data flows like Inception, Residual, and Dense blocks. It explores the design space based on the Synchronous Data Flow (SDF) model, accounting for platform resource constraints. Several optimizations have been proposed for fpgaConvNet, including a latency-driven design~\cite{venieris2017latency} and a heuristic method for pruning the architecture search space. Venieris \textit{et al.} \cite{venieris2018fpgaconvnet} later extended this framework to handle CNNs with irregular connectivity, treating design space exploration (DSE) as a Multi-Objective Optimization (MOO) problem to address throughput and latency requirements.

In contrast, Angel-Eye \cite{guo2017angel} and Snowflake \cite{gokhale2017snowflake} adopted a flexible computing engine design, allowing the reuse of hardware across different CNNs without recompilation. Snowflake further enhanced performance with double buffering to overlap computation and communication, hiding external memory latency.
f-CNNx \cite{venieris2018f} enabled mapping multiple CNNs onto a single FPGA, optimizing resource allocation and memory bandwidth to reduce contention and improve performance. It used a multi-objective cost function for efficient mapping.
FMM-X3D \cite{toupas2023fmm} optimized 3D-CNNs for human action recognition (HAR) on FPGAs, while HARFLOW3D \cite{toupas2023harflow3d} introduced runtime parameter reconfiguration to avoid full bitstream reconfiguration, improving latency. 
fpgaHART \cite{toupas2023fpgahart} extended the SDF model to manage irregular blocks with branching structures in 3D-CNN HAR models.
Pflow \cite{wan2024pflow} decoupled hardware details with an \textit{Overlay} library and employs adaptive scheduling and operator fusion to maximize hardware utilization, enabling end-to-end acceleration from application to hardware.

\begin{table*}[t]

\centering

\caption{Summary and Comparison of Existing CNN Acceleration Toolflows}

    \renewcommand\arraystretch{2.5}
    
    \resizebox{0.95\textwidth}{!}
    {
    \begin{tabular}{|c|m{2.5cm}<{\centering}|m{2.5cm}<{\centering}|m{2.5cm}<{\centering}|m{3cm}<{\centering}|m{3cm}<{\centering}|m{3cm}<{\centering}|}
    \hline
        \textbf {Ref.} &   \textbf {Toolflow Name} & \textbf{Type}& \textbf {Device} & \textbf {Throughput (Gop/s)} & \textbf {Energy EFF. (Gop/s/W)} & \textbf {Resource EFF. (GOp/s/DSP)}
        \\ \hline
        
        \cite{venieris2016fpgaconvnet} & fpgaConvNet & Streaming Structure &   Xilinx XC7Z020 &  \makecell[l]{ $\bullet$ 12.73 } & \makecell[l]{ $\bullet$ 7.27 }  & -
    \\ \hline   
        \cite{guo2017angel} & Angel-Eye & Single Engine & 
       Xilinx XC7Z045/XC7Z020 &  \makecell[l]{ $\bullet$ XC7Z045: 137 \\  $\bullet$ XC7Z020: 84.3 } & \makecell[l]{ $\bullet$ XC7Z045: 14.2 \\ $\bullet$ XC7Z020: 24.1  } & -
    \\ \hline   
        \cite{gokhale2017snowflake} & Snowflake & Single Engine & 
        Xilinx XC7Z045 & \makecell[l]{ $\bullet$ GoogLeNet: 116.5} & \makecell[l]{ $\bullet$ GoogLeNet: 12.3} & -
        \\ \hline   
        
        \cite{venieris2017latency} & fpgaConvNet & Streaming Structure &   
         Xilinx XC7Z045 & \makecell[l]{ $\bullet$ AlexNet: 134.10 \\ $\bullet$ VGG16: 123.12}   & - & \makecell[l]{ $\bullet$ AlexNet: 0.18 \\  $\bullet$  VGG16: 0.14} 
    \\ \hline 
    
        \cite{venieris2018fpgaconvnet} & fpgaConvNet & Streaming Structure & Xilinx XC7Z045 &  \makecell[l]{ $\bullet$ AlexNet: 197.40 \\ $\bullet$ VGG16: 155.81 \\ $\bullet$ GoogLeNet: 165.30 \\  $\bullet$ ResNet-152: 188.18 \\ $\bullet$ DenseNet-161: 155.57} & 
 \makecell[l]{ $\bullet$ AlexNet: 49.35 \\  $\bullet$ VGG16: 38.95  \\ $\bullet$ GoogLeNet: 41.32 \\ $\bullet$ ResNet-152: 47.04  \\ $\bullet$ DenseNet-161: 38.9 } &  \makecell[l]{ $\bullet$ AlexNet: 0.22 \\  $\bullet$ VGG16: 0.17 \\ $\bullet$ GoogLeNet: 0.184 \\ $\bullet$ ResNet-152: 0.209 \\ $\bullet$ DenseNet-161: 0.173 }
    \\ \hline     
    
        \cite{venieris2018f} & f-CNNx & Streaming Structure & 
        Xilinx XC7Z045& \makecell[l]{ $\bullet$ VGG16: 75.00}  &\makecell[l]{ $\bullet$ VGG16 : 18.75 }& -
    \\ \hline   
    
        \cite{toupas2023fmm} & FMM-X3D & Streaming Structure & 
        Xilinx ZCU102& \makecell[l]{ $\bullet$ X3D-M: 119.83 }  & - & \makecell[l]{ $\bullet$ X3D-M: 0.055} 
    \\ \hline  
    
        \cite{toupas2023harflow3d} & HARFLOW3D & Streaming Structure & 
        Xilinx ZCU102/VC709 & \makecell[l]{
        ZCU102:\\
        $\bullet$\quad C3D: 393.37 \\ $\bullet$\quad Slowonly: 177.05 \\ 
        $\bullet$\quad R(2+1)D-18: 173.91 \\  
        $\bullet$\quad (R(2+1)D-34): 184.29 \\ 
        $\bullet$\quad X3D-M: 43.78 \\
        VC709:\\
        $\bullet$\quad C3D: 424.14 \\ $\bullet$\quad Slowonly: 229.01 \\ 
        $\bullet$\quad R(2+1)D-18: 185.13 \\  
        $\bullet$\quad (R(2+1)D-34: 206.39 \\ 
        $\bullet$\quad X3D-M: 56.14} &-& \makecell[l]{
        ZCU102:\\
        $\bullet$\quad C3D: 0.156 \\ $\bullet$\quad Slowonly: 0.07 \\ 
        $\bullet$\quad R(2+1)D-18: 0.069 \\  
        $\bullet$\quad (R(2+1)D-34: 0.073 \\ 
        $\bullet$\quad X3D-M: 0.017 \\
        VC709:\\
        $\bullet$\quad C3D: 0.117 \\ $\bullet$\quad Slowonly: 0.063 \\ 
        $\bullet$\quad R(2+1)D-18: 0.051 \\  
        $\bullet$\quad (R(2+1)D-34: 0.057 \\ 
        $\bullet$\quad X3D-M: 0.015}
    \\ \hline  
    
        \cite{toupas2023fpgahart} & fpgaHART & Streaming Structure & 
        Xilinx ZCU102 & \makecell[l]{
        $\bullet$\quad C3D: 130.84 \\ $\bullet$\quad Slowonly: 144.44 \\ 
        $\bullet$\quad R(2+1)D-18: 39.59 \\  
        $\bullet$\quad (R(2+1)D-34: 34.26 \\ 
        $\bullet$\quad X3D-M: 85.96} & -& \makecell[l]{
        $\bullet$ C3D: 0.052 \\ $\bullet$ Slowonly: 0.057 \\ 
        $\bullet$ R(2+1)D-18: 0.015 \\  
        $\bullet$ (R(2+1)D-34: 0.013 \\ 
        $\bullet$ X3D-M: 0.034}
    \\ \hline   
    \cite{wan2024pflow} & Pflow & Streaming Structure & 
    Xilinx XCZU3EG/XCVU13P & \makecell[l]{$\bullet$ XCZU3EG: 272.64\\
    $\bullet$ XCVU13P: 3686.4 }& \makecell[l]{
    $\bullet$ XCZU3EG: 46.5 \\
    $\bullet$ XCVU13P: 59.4 }& \makecell[l]{$\bullet$ XCZU3EG: 0.71\\
    $\bullet$ XCVU13P: 0.6}
    \\ \hline
    \end{tabular}
    }
  \vspace{-1ex} 
    \label{tab:hh}
\end{table*}

\subsection{Design Space Exploration}

Design Space Exploration (DSE) involves systematically optimizing design parameters within set constraints. To maximize FPGA capabilities, factors such as network architecture, hardware resources, memory bandwidth, and latency must be considered. Optimizing these parameters enhances CNN inference while making full use of FPGA hardware features. Table \ref{tab:dse_method} compares various DSE methods, including brute-force and heuristic approaches.

\subsubsection{\textbf{Brute-Force Method}}
The Brute-Force  method \cite{montgomerie2022samo} exhaustively searches all possible solutions to find the global optimum. While it guarantees the best result, it is computationally expensive and time-consuming, making it impractical for large-scale optimizations.

\subsubsection{\textbf{Simulated Annealing}}
Simulated Annealing (SA) \cite{montgomerie2022samo, venieris2018fpgaconvnet} uses probabilistic jumping to explore the solution space, gradually lowering the search temperature to converge to the global optimum. While it prevents local optima, its convergence is slow, making it less efficient for large-scale searches.

\subsubsection{\textbf{Rule-Based Method}}
The Rule-Based method \cite{montgomerie2022samo} uses a deterministic approach, optimizing variables starting from a minimum resource state. It is faster than SA but may be limited by memory bandwidth constraints, impacting performance.

\subsubsection{\textbf{Genetic Algorithm}}

The Genetic Algorithm (GA) \cite{shim2014adaptive, yu2020software} simulates natural selection through selection, crossover, and mutation. It iteratively improves solutions based on fitness functions. GA is highly parallelizable and effective for global search but suffers from slow convergence and the risk of premature local optima.

 \subsubsection{\textbf{Other Methods}}
Biggs \textit{et al.} \cite{biggs2023atheena} used probability profiles to optimize network phases based on throughput and area trade-offs. \cite{liu2018optimizing} applied Fmincon for optimizing CNN accelerators, although its results can vary with initial conditions. \cite{xie2024design} introduced a Genetic Simulated Annealing (GSA) method, combining the strengths of GA and SA to explore a broader solution space with reduced computational time and fewer parameter dependencies.

\begin{table*}[t]

\centering

\caption{Comparison of Existing DSE Methods}

    \renewcommand\arraystretch{2}   
    \resizebox{0.95\textwidth}{!}   
    {
    
    \begin{tabular}{|m{3cm}<{\centering}|m{2.5cm}<{\centering}|m{2.5cm}<{\centering}|m{2.5cm}<{\centering}|m{2.5cm}<{\centering}|m{2.5cm}<{\centering}|m{2.5cm}<{\centering}|}
    \hline
        \textbf {DSE Method} & \textbf{Local Optimum} & \textbf {Large Parameter Space} & \textbf{Real-Time}&  \textbf {Low Complexity}  & \textbf{Low Parameter Dependency}  & \textbf{Memory Bandwidth Limitations}
    \\ \hline
        \textbf {BruteForce} &   $\times$ & $\times$ & $\times$  & $\checkmark $ & $\checkmark $   & $\times$
        \\ 
    \hline
        \textbf {Simulated Annealing} &   $\times$  & $\checkmark $  & $\checkmark $ & $\checkmark $ & $\checkmark $ & $\times$
        \\ 
    \hline
        \textbf {Rule-Based Method} &    $\times$ & $\checkmark $  & $\checkmark $ & $\times$ & $\checkmark $ & $\checkmark $
        \\ 
    \hline
        \textbf {Genetic Algorithm} & $\checkmark $  &$\checkmark $ & $\checkmark $ &$\checkmark $ & $\times$ & $\times$
        \\ 
    \hline
    \end{tabular}
    }
  \vspace{-1ex} 
    \label{tab:dse_method}
\end{table*}

\subsection{Performance and Resource Modeling}

The DSE process typically uses two approximate models: resource and performance models. The resource model estimates the hardware requirements for a given architecture on the target FPGA, while the performance model predicts system performance (e.g., latency) based on hardware configuration and algorithm characteristics. These models replace the time-consuming task of running CNNs on actual hardware by providing estimates based on hardware parameters. Analytical prediction formulas are often used for simplicity and integration into DSE optimization loops, compared to more time-intensive full simulations like ModelSim \cite{liu2019towards}.

\subsubsection{\textbf{Performance Modeling}}

Liu \textit{et al.} \cite{liu2021toward} combined Gaussian Process Regression (GPR) \cite{williams1995gaussian} with analytical formulas~\cite{liu2018optimizing} to estimate CNN latency on accelerators. Makrani \textit{et al.}~\cite{makrani2019pyramid} introduced the Pyramid framework, using the Minerva tool and ensemble learning to improve throughput prediction accuracy, achieving over 95\% accuracy. The Intel FPGA Power and Thermal Tool \cite{intel_power} estimated power consumption based on FPGA model, frequency, and resource usage, though discrepancies may occur. PowerGear \cite{lin2022powergear} used a Heterogeneous Edge-Centered Graph Neural Network (HEC-GNN) to predict power consumption efficiently.

\subsubsection{\textbf{Resource Modeling}}

Dai \textit{et al.} \cite{dai2018fast} developed a dataset with over 1,300 samples to predict resource usage and timing during HLS with machine learning models. Koeplinger \textit{et al.} \cite{koeplinger2016automatic} introduced DHDL, a hardware representation method that used parameterized templates to capture locality and parallelism in designs. Lee \textit{et al.} \cite{lee2015dynamic} presented a framework for fast, accurate resource estimation by combining high-level hardware descriptions with low-level performance models, offering cycle-level timing and power estimates with high simulation speed.



\section{Discussion and Conclusion}
\label{sec:discuss}

\subsection{Discussion}

\subsubsection*{Insights and optimization}
Based on the analysis presented in previous sections, several key challenges in CNN accelerator design have been identified, along with corresponding optimization strategies to address these challenges. These insights can guide future research and development efforts in enhancing the performance and efficiency of CNN hardware accelerators.


\textbf {1) Clock Efficiency Optimization.} 
One of the primary concerns in FPGA-based CNN accelerators is maximizing clock efficiency. FPGAs inherently offer fine-grained control over clock frequencies, which can be leveraged to optimize resource utilization. Dynamic frequency scaling, where high clock frequencies are assigned to computation-heavy tasks (e.g., convolution layers) and lower frequencies to less intensive operations, helps minimize energy consumption without sacrificing performance. Additionally, optimizing clock network design and clock gating reduces unnecessary switching activity by deactivating idle modules, contributing further to power savings. Pipelining techniques are also employed to reduce critical path delays, ensuring that the FPGA operates efficiently at higher frequencies while maintaining stable synchronization. In multi-clock domain systems, asynchronous FIFOs or synchronous bridges are utilized to prevent metastability, enhancing stability and performance across different clock domains.

\textbf {2) Power Efficiency Optimization.}
Power consumption is a key challenge, particularly in edge applications where energy efficiency is critical. FPGA-based CNN accelerators employ several techniques to reduce power usage. Weight pruning and feature map sparsification lower the computational load, cutting dynamic power. Low-bit-width quantization reduces both memory usage and computation time, further conserving power. On-chip storage and computation fusion minimize power-hungry off-chip memory accesses. Power management methods like dynamic voltage and frequency scaling (DVFS) adjust power and frequency based on workload, while clock gating shuts down idle modules. Architectural strategies such as deep pipelining, parallelism, and reconfigurable units improve both computation and energy efficiency. Choosing low-power FPGA devices and optimizing architecture to remove non-essential resources helps minimize static power, making FPGAs ideal for low-power, real-time applications.



\subsubsection*{Future directions}
Several promising directions for future research in CNN hardware acceleration can significantly enhance the performance, efficiency, and versatility of hardware accelerators, especially in the context of emerging applications in AI, computer vision, and edge computing. The following are key areas for continued exploration:


\textbf {1) Heterogeneous Computing Architectures.} 
Integrating multiple computing units (CPU, GPU, FPGA, ASIC) on a single chip could significantly enhance resource utilization and energy efficiency for CNN tasks. Key technologies like optimized Network-on-Chip (NoC) and intelligent task scheduling will drive this development, though challenges in hardware complexity and software support remain.


\textbf {2) AI-Driven Design Space Exploration (DSE).} 
As CNN models continue to grow in complexity, the need for efficient design space exploration (DSE) methods becomes increasingly important. Traditional DSE methods often require exhaustive searching of vast design spaces, which is time-consuming and computationally expensive. Future research should focus on integrating machine learning techniques, such as reinforcement learning or neural architecture search, into the DSE process. AI-driven DSE can dynamically explore design spaces based on performance metrics, such as throughput, latency, and power consumption, enabling the rapid discovery of optimal hardware configurations. By automating this process, researchers can significantly reduce design time and enhance the ability to tailor hardware accelerators to specific CNN models and tasks.



\textbf {3) Scalable and Adaptive CNN Hardware Accelerators.} 
As CNN models continue to grow in size and complexity, accelerators must be able to scale efficiently. Scalable hardware accelerators capable of dynamically adapting to different CNN architectures and input sizes are essential. Future research could focus on developing adaptive FPGA architectures that allow for dynamic reconfiguration based on the model being run. These accelerators would dynamically adjust resources (e.g., memory, processing units) and parallelism levels to accommodate the requirements of different layers in the CNN. Such adaptive systems could support a broader range of CNN applications, from small, lightweight models to large, deep networks, without requiring a complete redesign of the hardware.

\subsection{Conclusion}

CNNs are integral to AI-driven applications, but the challenge remains to achieve low-cost, low-latency, and high-performance solutions. This paper reviewed various hardware acceleration platforms, comparing their strengths and weaknesses. We also discussed key evaluation metrics for both software and hardware, such as accuracy, latency, and throughput. By examining popular CNN acceleration techniques, we identified the advantages and limitations of each approach. Our evaluation emphasized the importance of computational efficiency and dynamic parallel computing in optimizing CNN hardware accelerators.
As the field continues to evolve, future research should focus on dynamic parallel computing and heterogeneous on-chip computing for CNN acceleration. With increasing design complexity, effective design space exploration will also become crucial.

\section*{Acknowledgement}{
The support of Scientific Research Foundation of Hunan Provincial Education Department (Key project 23A0087) and Changsha Municipal Natural Science Foundation (No. kq2502001) is gratefully acknowledged.
}

\bibliographystyle{IEEEtran}
\bibliography{bibliography} 

\begin{IEEEbiography}
[{\includegraphics[width=1in,height=1.25in,clip,keepaspectratio]{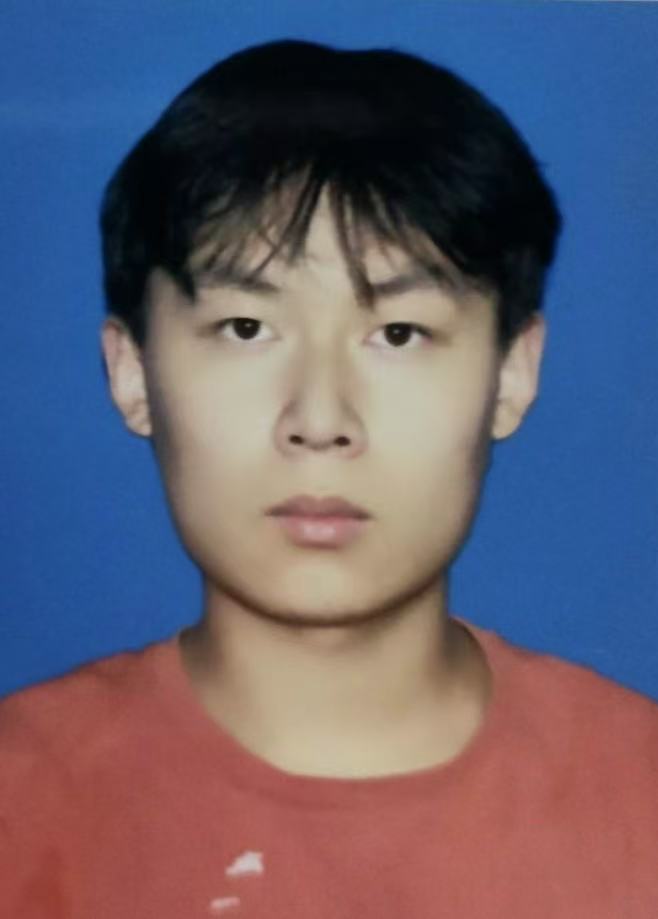}}]
{Junye Jiang}
received the B.Sc. degree from the School of Electrical and Information Engineering, Wuhan Institute of Technology, Wuhan, China, in 2023.
He is currently a M.Sc. student in the Computer Vision and High-Performance Computing group at Hunan Normal University.
His research interests include the hardware acceleration of convolutional neural networks (CNNs) and neural architecture design optimizations applied to computer vision tasks. 
\end{IEEEbiography}

\begin{IEEEbiography}
[{\includegraphics[width=1in,height=1.25in,clip,keepaspectratio]{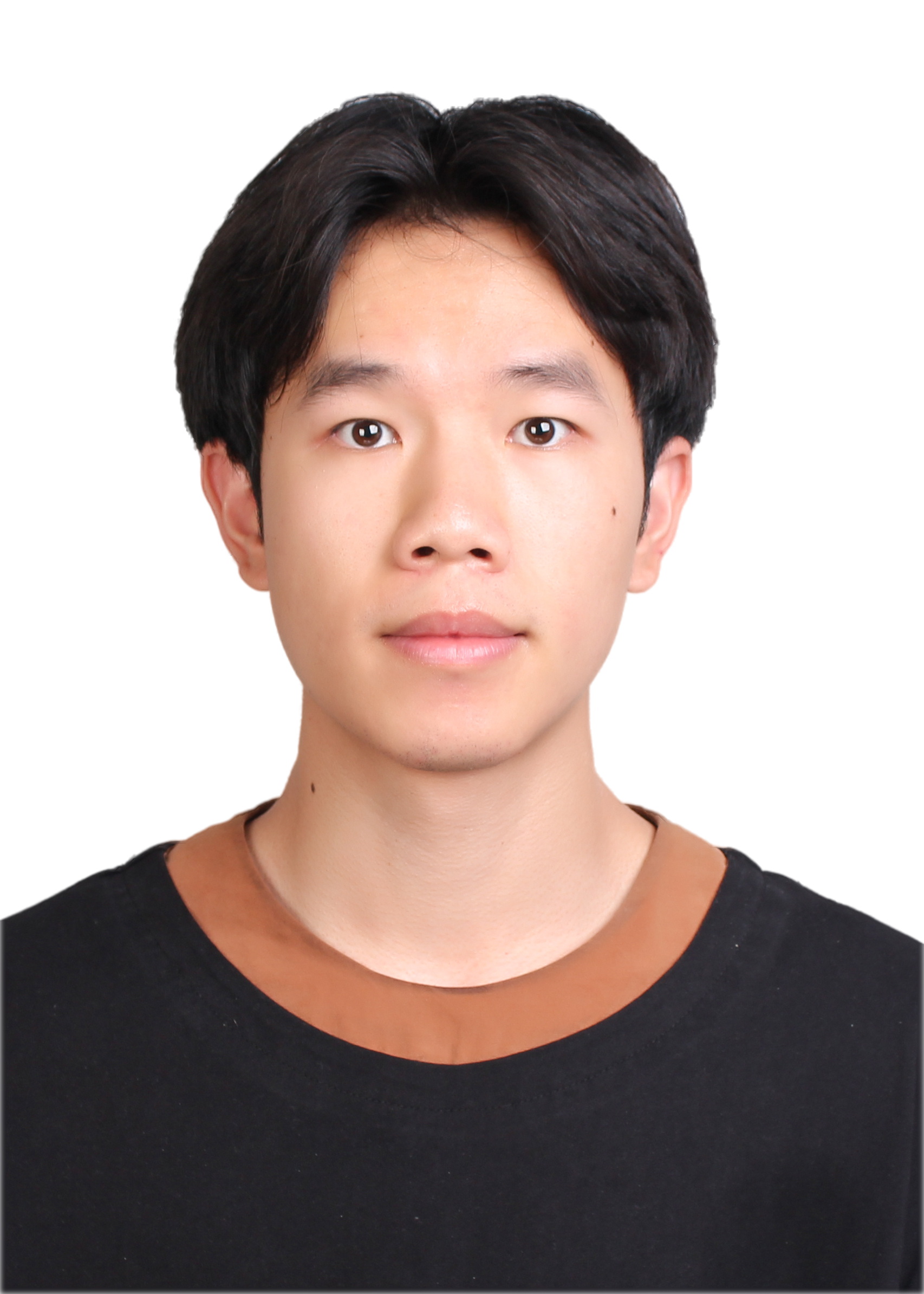}}]
{Yaan Zhou}
received the B.Sc. degree in Electronic Information Engineering of Central South University of Forestry and Technology, Changsha, China, in 2021. He is currently a M.Sc. student in the Computer Vision and High-Performance Computing group at Hunan Normal University. His current research includes the hardware acceleration of convolutional neural networks (CNNs). 
\end{IEEEbiography}

\begin{IEEEbiography}[{\includegraphics[width=1in,height=1.25in,clip,keepaspectratio]{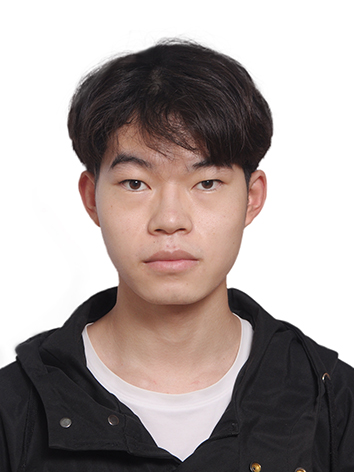}}]
{Yuanhao Gong} received the B.S. degree in Electronic Information Science and Technology from Hunan Normal University, Changsha, China, in 2023. He is currently pursuing the M.S. degree in Electronic Science and Technology with Hunan Normal University. His research interests mainly focus on particle filter methods and efficient hardware architectures for particle filters.
\end{IEEEbiography}

\begin{IEEEbiography}[{\includegraphics[width=1in,height=1.25in,clip,keepaspectratio]{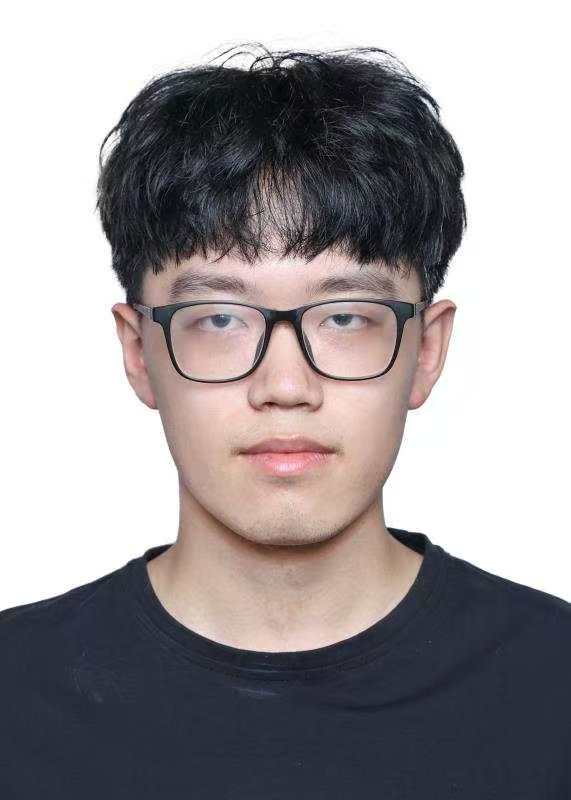}}]
{Haoxuan Yuan} received the B.S. degree in Electronic Information Science and Technology from Hunan Normal University, Changsha, China, in 2024. He is currently a M.Sc. student in the Computer Vision and High-Performance Computing group at Hunan Normal University. His research interests include the reconfigurable acceleration of artificial intelligence algorithms. 
\end{IEEEbiography}

\begin{IEEEbiography}
[{\includegraphics[width=1in,height=1.25in,clip,keepaspectratio]{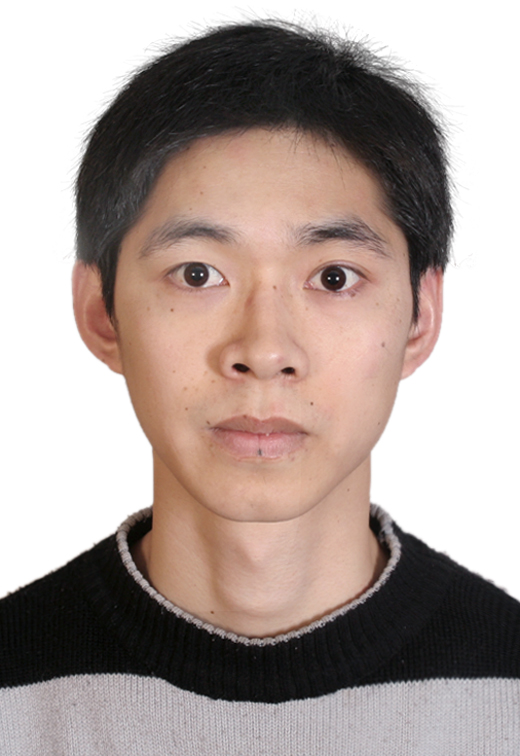}}]
{Shuanglong Liu}
received the B.Sc. and M.Sc. degrees from the Department of Electronic Engineering, Tsinghua University, Beijing, China, in 2010 and 2013 respectively, and Ph.D. degree in Electric Engineering from Imperial
College London, London, U.K, in 2017.
From 2017 to 2020, he was a Research Associate with the Department of Computing, Imperial College London. 
He is currently a Distinguished Professor in the School of Physics and Electronics, Hunan Normal University, Changsha, China. He has published
over 40 research papers in peer-referred journals
and international conferences. His current research interests
include reconfigurable and high performance
computing for Convolutional Neural Networks (CNNs) and statistical inference problems.
\end{IEEEbiography}

\end{document}